\documentclass[11pt]{article}

\usepackage[final]{acl}

\usepackage{times}
\usepackage{latexsym}

\usepackage[T1]{fontenc}

\usepackage[utf8]{inputenc}

\usepackage{microtype}

\usepackage{inconsolata}

\usepackage{graphicx}
\usepackage{amsmath} 
\usepackage{booktabs} 
\usepackage{enumitem}
\usepackage{times}  
\usepackage{helvet}  
\usepackage{courier}  
\usepackage{graphicx} 

\usepackage{algorithmic}
\usepackage{makecell}
\usepackage{booktabs}
\usepackage{multirow}
\usepackage{xcolor}
\usepackage{colortbl}
\usepackage{tcolorbox}
\usepackage[linesnumbered,ruled,vlined]{algorithm2e}
\usepackage{float}

\title{VRPO: Rethinking Value Modeling for Robust RL under  Noisy \\ Supervision in LLM Post-Training }
\author{
  \textbf{Dingwei Zhu$^{1}$}\thanks{\ \ Equal contribution.},
  \textbf{Shihan Dou$^{1*}$},
  \textbf{Zhiheng Xi$^{1*}$},
  \textbf{Senjie Jin$^1$},
  \textbf{Guoqiang Zhang$^1$}, \\
  \textbf{Jiazheng Zhang$^1$},
  \textbf{Junjie Ye$^1$},
  \textbf{Mingxu Chai$^1$},
  \textbf{Enyu Zhou$^1$},
  \textbf{Ming Zhang$^1$}, \\
  \textbf{Yuhui Wang$^1$},
  \textbf{Caishuang Huang$^1$},
  \textbf{Chenhao Huang$^1$},
  \textbf{Yunke Zhang$^2$},
  \textbf{Yuran Wang$^2$}, \\
  \textbf{Tao Gui$^{1}$},
  \textbf{Qi Zhang$^1$},
  \textbf{Xipeng Qiu$^1$},
  \textbf{Xuanjing Huang$^{1,3,4}$}\thanks{\ \ Corresponding author.}\\
  \vspace{0.1cm} 
  \small $^1$College of Computer Science and Artificial Intelligence, Fudan University \\
  \small $^2$Honor Device Co., Ltd \quad 
  \small $^3$Institute of Trustworthy Embodied AI, Fudan University \\
  \small $^4$Shanghai Key Laboratory of Multimodal Embodied AI \\
  \texttt{\small dwzhu25@m.fudan.edu.cn, xjhuang@fudan.edu.cn}
}

\begin{document}
\maketitle
\begin{abstract}
Reinforcement Learning (RL) in real-world environments often suffers from ambiguous or incomplete reward supervision, which undermines policy stability and generalization. Such noise may cause models to ignore key information or even collapse in advantage estimation. We find that a strong value model is essential for absorbing unstable signals and producing reliable advantages, offering denser and more robust supervision than the reward model. 
To better optimize noisy supervision, we propose VRPO, a framework that enhances value modeling for robust RL in LLM post-training. VRPO integrates (1) auxiliary losses guided by entropy and perplexity from a frozen language model, and (2) a variational information bottleneck, enabling the value model to filter noise and capture key words. This design allows the value model to correct noise rewards and generate more reliable advantage estimates, transforming it from a passive predictor into an active noise regulator. 
Experiments on multi-turn dialogue, math reasoning, and science QA with both rule-based and model-based rewards show that VRPO consistently outperforms baselines such as PPO and GRPO. Our work  highlight the central role of the value model in Robust RL and provide a principled and practical approach to policy optimization under noisy supervision.
\end{abstract}

\section{Introduction}
\begin{figure*}[h]
\centering
\includegraphics[width=1\linewidth]{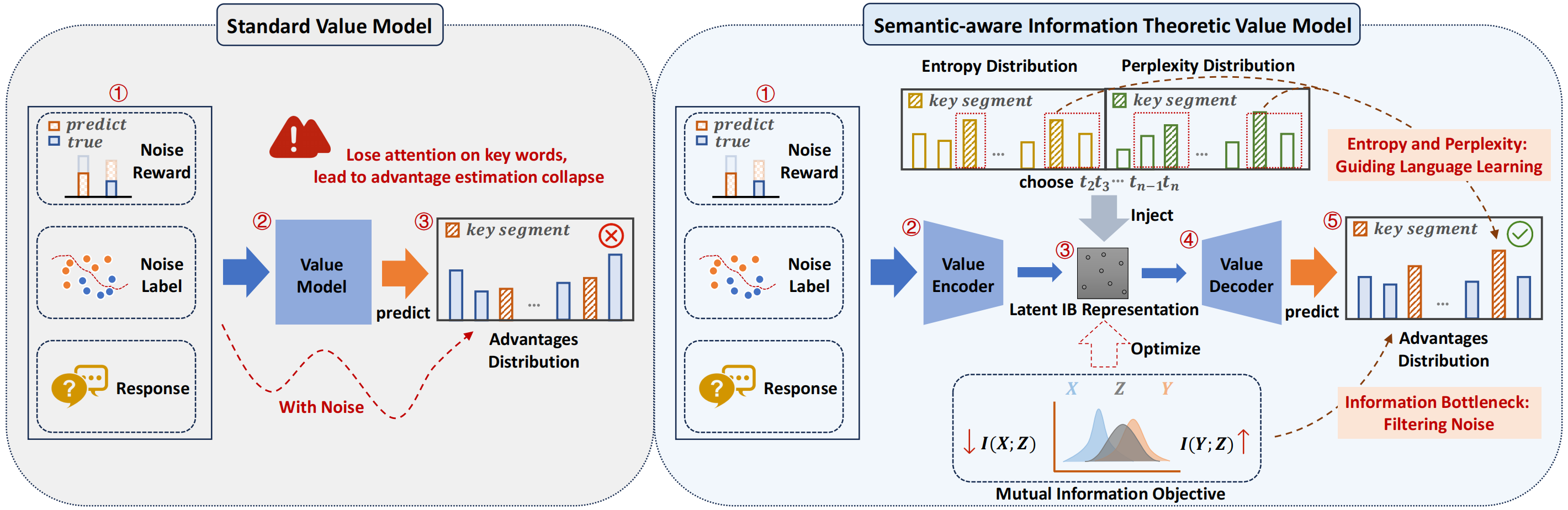}
\caption{Comparison between the Standard Value Model and our Semantic-aware Information Theoretic Value Model.
The standard model fits noisy rewards and labels, leading to unstable advantage estimates. VRPO enhances robustness by guiding learning with entropy and perplexity signals, and filtering noise via a variational information bottleneck, yielding more reliable value predictions and improved PPO stability.}
\label{fig:all}
\end{figure*}

Reinforcement Learning has achieved remarkable success across a wide range of applications~\cite{2022Sci...378..990P,2022arXiv220608686C,Bellemare2020AutonomousNO,2020arXiv201009776Z,2024arXiv240313263Y,lin2026mmdocr1trainingagentslong}. However, deploying RL in real-world scenarios often involves noisy or imperfect supervision, particularly when optimization depends on human feedback or learned reward models. This challenge is especially evident in LLM Post-Training RL and preference-based Reinforcement Learning from Human Feedback (RLHF)~\cite{2023arXiv230715217C,2025arXiv250510597Z}, where reward signals are approximate and not directly derived from ground-truth annotations, such as those generated by Generalized Advantage Estimation (GAE)~\cite{gae}.

To address this issue, recent works have proposed robust RL training methods by either denoising reward models~\cite{2025arXiv250318991C,InfoRM} or filtering corrupted data before policy updates~\cite{RIME,2024arXiv240812109W}. However, these methods implicitly assume that reward errors can be corrected during training, which often does not hold in practice: training reward  are frequently ambiguous, sparse, or fundamentally unreliable~\cite{2024arXiv240103857P}. As a result, inaccurate reward signals can propagate throughout RL training, leading to instability in textual perception, loss of critical information during advantage estimation, and ultimately degraded policy optimization performance and convergence stability.

While noisy reward modeling has received considerable attention, the value model, which provides denser supervision and plays a central role in training, has been largely overlooked as a potential denoising mechanism. In this paper, we adopt an information-theoretic perspective~\cite{hung2023valuebiasedmaximumlikelihoodestimation} and propose an alternative approach: rather than relying solely on reward estimation or data filtering, we enhance RL robustness by directly optimizing the value model to absorb uncertainty and stabilize training. This perspective highlights the value model as a critical component for guiding language perception under noisy supervision.

We introduce \textbf{VRPO} (Value Model Boosting for Robust Policy Optimization), a novel framework that integrates a noise-resilient and semantically aware value model into RL training. VRPO combines two key innovations: (1) an auxiliary loss guided by entropy and perplexity from a frozen language model, and (2) a variational information bottleneck (IB) architecture for robust value modeling. The IB mechanism constrains irrelevant information flow, enhancing tolerance to noise, while language signals from the frozen LM guide model to capture key words under noisy training. These auxiliary objectives further align the model’s internal feature space with the semantic space of language, effectively suppressing spurious noise while retaining task-relevant information. Moreover, VRPO alleviates overfitting~\cite{2022arXiv221010760G} and mitigates contextual noise~\cite{MINER}, thereby improving training robustness.

We evaluate VRPO across multiple tasks, including multi-turn dialogue, mathematical reasoning, and scientific question answering, under both rule-based and model-based noisy reward settings. Experimental results show that VRPO consistently outperforms baselines such as PPO and GRPO in noisy supervision scenarios. By incorporating information-theoretic regularization and semantic guidance into RL training, VRPO transforms the value model from a passive estimator into an active, noise-aware regulator. Through context-appropriate advantage estimation and robust key-signal extraction, our method significantly enhances stability and generalization in complex environments.

\noindent\textbf{Our contributions are as follows:}
\begin{itemize}
    \item We propose entropy and perplexity-based auxiliary losses that mitigate advantage collapse from noisy rewards improving the semantic robustness of the value model. 
    \item We introduce a variational information bottleneck structure that constrains irrelevant information flow.
    \item We empirically demonstrate that VRPO improves RL robustness and generalization in noisy RLHF and RL scenarios, offering a principled and scalable solution for real-world RL applications.
\end{itemize}

\section{Related Work}
Robust RL methods primarily focus on improving reward models or filtering noisy data. Reward-oriented approaches build noise-resistant discriminators via conservative gradients~\cite{ROPO}, ensemble models \cite{Secrets}, or uncertainty-aware losses \cite{2024arXiv240707880W}. For example, data-centric methods like RIME \cite{RIME} use pre-trained denoisers with KL bounds or apply reward consensus to remove inconsistent preferences \cite{Secrets}. Our work shifts robustness efforts from reward correction to value model compensation. Through value model optimization, we partially absorb and filter out the noise propagation throughout the training process.

The Information Bottleneck (IB) framework offers a principled way to learn compact \cite{2018arXiv180210399D,2019arXiv190110902G,2019arXiv191201603H,2018arXiv181106521I}, robust representations by balancing compression and prediction \cite{MINER} through variational approximations \cite{JMLR:v26:24-0204}.   In RL, IB has enhanced policy or reward robustness under noisy supervision \cite{InfoRM} but has rarely been applied to value models.   We fill this gap by integrating IB regularization into the value function, enabling it to absorb noise during RL training.

\section{Method}
\subsection{Motivation}
In standard Proximal Policy Optimization (PPO), policy updates depend on the joint contribution of reward signals and value estimates via advantage computation. The core optimization objective is:

\begin{equation}\label{lppo}
\small
\mathcal{L}_{\text{PPO}} = \mathbf{E}_t\left[ \min\left( r_t(\theta) \hat{A}_t, \text{clip}(r_t(\theta), 1 - \epsilon, 1 + \epsilon)\hat{A}_t \right) \right],
\end{equation}
where the advantage $\hat{A}_t$ is typically computed through Generalized Advantage Estimation (GAE):

\begin{equation}\label{gae}
\small
\hat{A}_t = \delta_t + (\gamma \lambda)\delta_{t+1} + (\gamma \lambda)^2\delta_{t+2} + \cdots,
\end{equation}
\begin{equation}\label{ga}
\small
\delta_t = r_t + \gamma V(s_{t+1}) - V(s_t).
\end{equation}

This definition highlights the strong coupling between reward signals $r_t$ and the value function $V(s)$. When $r_t$ is noisy, its error propagates through $\hat{A}_t$, leading to unstable training and degraded final policy performance.

\begin{figure*}[h]
\centering
\includegraphics[width=1\linewidth]{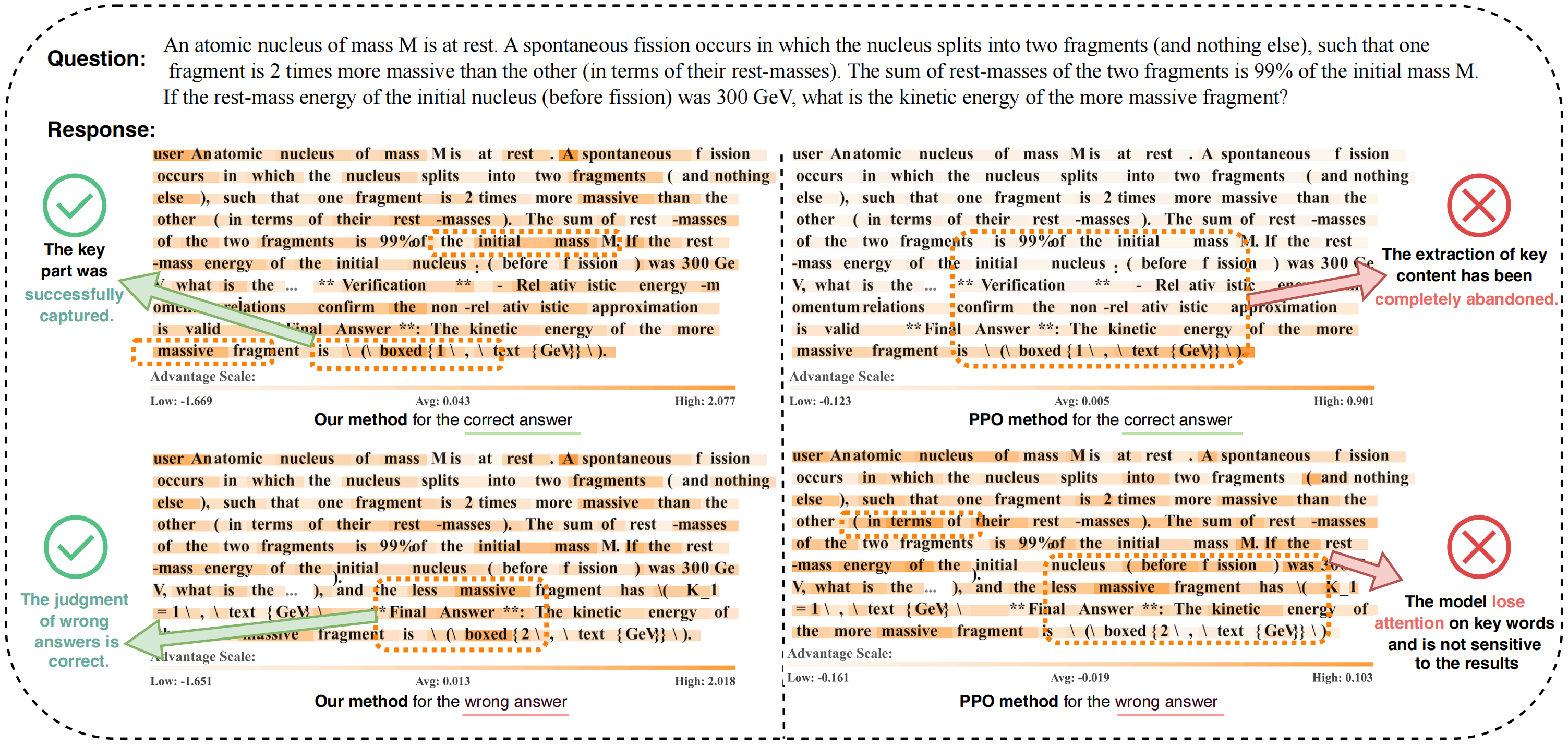}
\caption{Token-level advantage estimation for the same response across different methods. Our method exhibits sharper \textbf{focus on critical reasoning steps}.}
\label{fig:adv-vis}
\end{figure*}

We observe that under noisy supervision, the model often \textbf{misallocates advantage values} and fails to emphasize semantically important tokens. As illustrated in Figure~\ref{fig:adv-vis} (right), for correct answers, the model tends to concentrate higher advantages toward the sequence end, resulting in \textbf{length hacking}~\cite{chaudhari2024rlhfdecipheredcriticalanalysis,laidlaw2025correlatedproxiesnewdefinition}, where longer responses are rewarded regardless of quality. For incorrect answers, advantages are incorrectly assigned to irrelevant tokens. The root cause of such errors lies in biased reward signals: noisy $r_t$ induces distorted advantages under GAE.

To address this, we propose a complementary perspective: \textbf{enhancing the value model itself to compensate for biased advantage estimation under noisy rewards}. Specifically, we design value models guided by semantic signals from a frozen language model, ensuring robustness to linguistic noise and preserving attention to key tokens. Additionally, we incorporate an information bottleneck mechanism to constrain irrelevant information flow, thereby improving the value model’s noise tolerance and stabilizing training. As shown in Figures~\ref{fig:adv-vis} , this design not only preserves critical signals in advantage estimation but also suppresses fluctuations under irrelevant perturbations.

Compared with reward-only denoising approaches, our framework leverages denser value supervision to more effectively correct distorted advantages, leading to superior training stability and final policy performance.

\subsection{Information-Theoretic Value Modeling}

To mitigate reward noise propagation in PPO, the value model must extract return-relevant information while filtering task-irrelevant details. This capability is essential for robust training under ambiguous, sparse, or inconsistent reward signals.

We address this by reformulating value modeling from an information-theoretic perspective. Let the input be a random variable \( X \), the latent representation \( Z \), and the return \( Y \). Assuming \( Z \) follows a Gaussian distribution, we define:
\begin{equation}\label{vaeib}
\small
I_{\text{bottleneck}} = I(X; Z), \quad I_{\text{value}} = I(Z; Y)
\end{equation}
Here, \( I_{\text{bottleneck}} \) measures information retained from the input, while \( I_{\text{value}} \) quantifies predictive information for returns. The objective is to learn a latent representation that maximizes return relevance while minimizing input redundancy:
\begin{equation}\label{vaej}
\small
\max J(\theta) = I(Z; Y) - \beta I(X; Z)
\end{equation}
where \( \beta \) controls the trade-off between compression and relevance, and \( \theta \) denotes model parameters.

Since mutual information is intractable in high-dimensional settings, we optimize a variational lower bound using training data \( D = \{(x_i, y_i)\}_{i=1}^N \):
\begin{equation}\label{vaej1}
\small
J(\phi, \psi) \geq J_{\text{VLB}}(\phi, \psi) = \mathbf{E}_{(x, y) \sim D}[J_{\text{value}} - \beta J_{\text{bottleneck}}]
\end{equation}
\begin{equation}\label{vaejv}
\small
J_{\text{value}} = \mathbf{E}_{z \sim p_{\phi}(z | x)}[\log q_{\psi}(y | z)]
\end{equation}
\begin{equation}\label{vaejb}
\small
J_{\text{bottleneck}} = \text{KL}[p_{\phi}(z | x) \| r(z)]
\end{equation}
where \( p_{\phi}(z | x) \) is the variational encoder, \( q_{\psi}(y | z) \) the return predictor, and \( r(z) \) the standard Gaussian prior. Parameters \( \phi \) and \( \psi \) correspond to the encoder and decoder.

In practice, we model \( p_{\phi}(z | x) \) as a diagonal Gaussian:
\begin{equation}\label{vaez}
\small
z = f_{\phi}^{\mu}(x) + f_{\phi}^{\sigma}(x) \cdot \epsilon, \quad \epsilon \sim \mathcal{N}(0, I)
\end{equation}
and use a lightweight MLP \( q_{\psi} \) to predict returns from \( z \). The final training objective becomes:
\begin{equation}\label{vaelv}
\small
\begin{split}
\mathcal{L}_{\text{value}} = \mathbf{E}_{(x, y) \sim D}[-\log q_{\psi}(y | z) \\
+\beta \cdot \text{KL}(p_{\phi}(z | x) \| r(z))]
\end{split}
\end{equation}

This structure encourages the value model to learn compact, reward-relevant latent representations while suppressing irrelevant or noisy input signals.

\subsection{Enhancing Semantic Awareness for Noise-Resistant Value Modeling}
While the information bottleneck in Section 3.2 helps filter out irrelevant features, it does not directly address semantic misalignment. Under noisy reward supervision, the value model may still lose attention on key words, resulting in prediction errors. To mitigate this, we introduce a semantic-level regularization mechanism that enhances the alignment between the model’s feature space and the language space.

Concretely, we incorporate a frozen LM head into the value model to express its internal token-level understanding.   We then guide this understanding using auxiliary losses based on entropy and perplexity, encouraging semantic consistency with the actor model despite noisy rewards.
Let \( P_V(y_t | x) \) denote the token-level prediction distribution from the LM head. We define the auxiliary losses as:
\begin{equation}
\small
L_{\text{ent}} = \sum_{t \in T_{\text{ent}}} H[P_V(y_t | x)], \quad   \\
\end{equation}
\begin{equation}
\small
L_{\text{ppl}} = \sum_{t \in T_{\text{ppl}}} -\log P_V(y_t = y_t^* | x)
\end{equation}
where \( H[\cdot] \) is entropy, and \( T_{\text{ent}} \), \( T_{\text{ppl}} \) are dynamically selected token subsets with high entropy and perplexity respectively:
\begin{equation}
\small
T_{\text{ent}} = \{ t : H[P_V(y_t | x)] > \hat{T}_{\text{entropy}} \},
\end{equation}
\begin{equation}
\small
T_{\text{ppl}} = \{ t : -\log P_V(y_t = y_t^* | x) > \hat{T}_{\text{perplexity}} \}
\end{equation}
where $\hat{T}_{\text{entropy}}$ and $\hat{T}_{\text{perplexity}}$ represent the threshold of entropy and perplexity. The final semantic regularization loss is:
\begin{equation}
\small
L_{\text{semantic}} = \lambda_{\text{ent}} L_{\text{ent}} + \lambda_{\text{ppl}} L_{\text{ppl}}
\end{equation}
where $\lambda_{\text{ent}}$and $\lambda_{\text{ppl}}$ represent the weight of the loss.

This method provides the value model with stable, semantically meaningful supervision, helping it stay anchored to the input’s linguistic structure and resist reward noise. Combined with the information bottleneck from Section~3.2, it reinforces both semantic alignment and robustness.

\section{Experiments}
\subsection{Setup}

We evaluate VRPO in two distinct but complementary settings: \textbf{rule-based rewards} and \textbf{model-based rewards}, both designed to test robustness under noisy supervision.

\textbf{Rule-based reward setting} This scenario mainly \textbf{considers the noise in unsupervised learning}. We simulate  supervision by performing majority voting over multiple outputs from pretrained models. Two variants are considered: (i) \emph{training-time augmentation} where noisy pseudo-labels from the training set supervise RL training and (ii) \emph{test-time augmentation}, where noisy pseudo-labels from the test set supervise RL training. Models are cold-started via supervised fine-tuning and constrained to specific answer formats.The results of test-time augmentation can be found in the appendix.


\textbf{Model-based reward setting} We conduct experiments on multi-turn dialogue, using a separate reward model trained on annotated dialogue data. A policy model is optimized via RL under this noisy reward. 

\textbf{Baselines} In both settings, we compare VRPO with PPO and GRPO under identical initialization. In addition, we include comparisons with several strong robust reinforcement learning baselines, including Dr.GRPO, Reinforce++, KTAE\cite{sun2025ktaemodelfreealgorithmkeytokens}, and $\lambda$-GRPO\cite{wang2025lambdagrpounifyinggrpoframeworks}.
 All models are based on Qwen2.5 \cite{qwen25}, Qwen3 \cite{qwen3} or Llama3.1 variants, fine-tuned appropriately.

\begin{figure}[h]
\centering
\includegraphics[width=1 \linewidth]{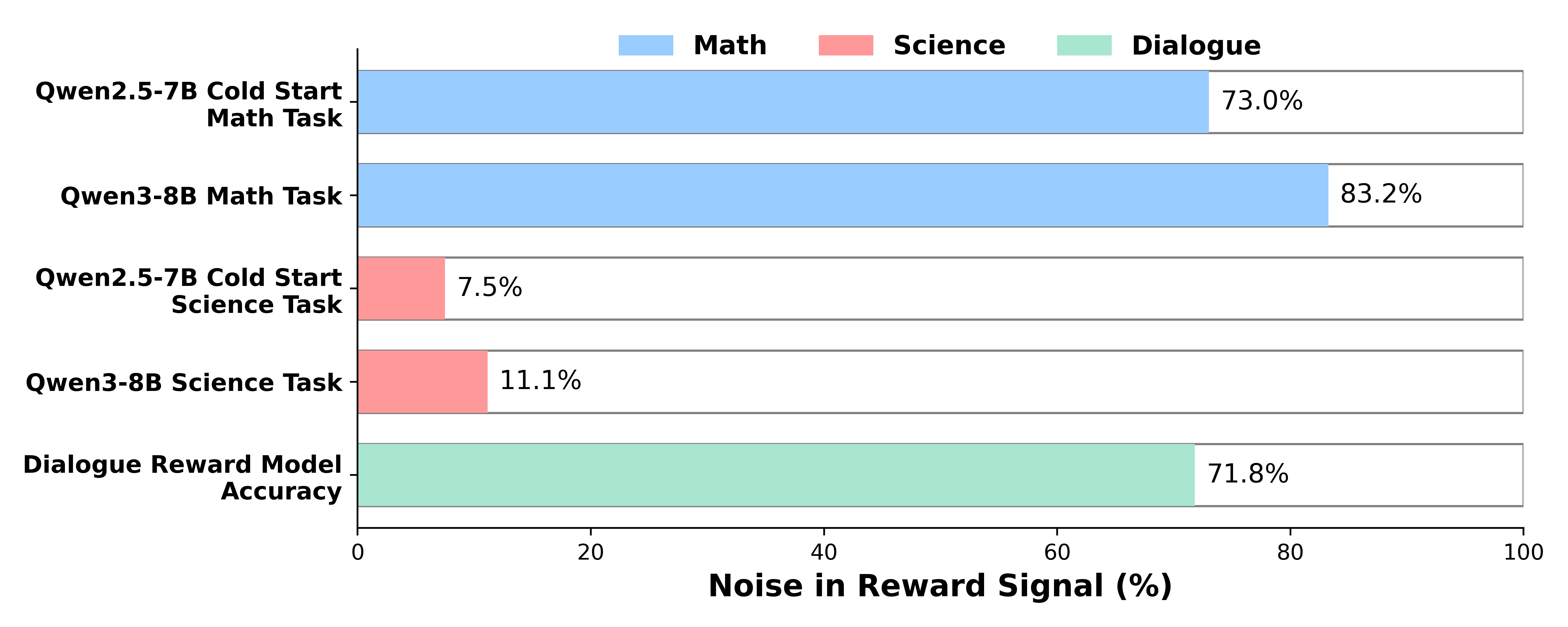}
\caption{Noise statistics in the various tasks. A significant portion of rewards contains inaccuracies.}
\label{fig:task-noise}
\end{figure}
\textbf{Training Noise} Noise is pervasive in RL training , especially under model-based feedback. Figure~\ref{fig:task-noise} illustrates the error rate in various reward signals. The noise in the rule-based reward from the majority vote generated data.

\textbf{Implementation Details} In the majority voting during test-time optimization, 32 samples are used, while 5 samples are used for voting during training-time optimization. The weights for the auxiliary losses (entropy and perplexity) are set to 0.5, making 80\% of the labels effective. The experiments are conducted on 8 NVIDIA A100 80GB GPUs.
For further dataset construction, training details, and model initialization, please refer to Appendix. 

\subsection{Evaluation Metrics}

To evaluate the effectiveness of our value model, we consider both task-level performance and value estimation capabilities across different task domains. 

\subsubsection{Task-Level Metrics}

 For mathematical reasoning and scientific knowledge tasks, we adopt accuracy as the primary metric to measure whether the model produces correct answers. For multi-turn dialogue tasks, we employ a tripartite evaluation framework that captures both task execution and communicative quality: (1) Task Completion Rate (TCR), assessing whether the model successfully fulfills the user-intended objective; (2) Ask Completion Rate (ACR), measuring whether detailed aspects of the user query are adequately addressed; and (3) Goal Completion Rate (GCR), evaluating the fluency, coherence, and appropriateness of the model's responses. This comprehensive evaluation allows us to examine not only whether the model achieves the intended outcomes, but also how effectively it aligns with human communication standards.

\subsubsection{Value Model Performance Metrics}

To directly evaluate the accuracy and robustness of the value model, we adopt the following criteria:

\paragraph{Explained Variance} We use the explained variance to measure how much of the empirical return is captured by the predicted value:
\begin{equation}\label{ev}
\small
\text{Explained Variance} = 1 - \frac{\text{Var}[\hat{R} - V]}{\text{Var}[\hat{R}]}
\end{equation}
where $\hat{R}$ denotes the empirical return and $V$ is the predicted value. A value near 1 indicates high explanatory power, while values near or below zero indicate poor or misleading predictions.

\paragraph{Prediction Error Evaluation}

To assess the value model’s prediction accuracy, we adopt Mean Squared Error (MSE) and Mean Absolute Error (MAE) between the predicted value \( V_\theta(s_t) \) and the reference target \( V_{\text{target}}(s_t) \):
\begin{equation}
\small
\mathcal{L}_{\text{MSE}} = \frac{1}{T} \sum_{t=0}^{T-1} \left( V_\theta(s_t) - V_{\text{target}}(s_t) \right)^2,
\end{equation}
\begin{equation}
\small
\mathcal{L}_{\text{MAE}} = \frac{1}{T} \sum_{t=0}^{T-1} \left| V_\theta(s_t) - V_{\text{target}}(s_t) \right|
\end{equation}

Here, \( T \) denotes the trajectory length, and \( V_{\text{target}}(s_t) \) is computed via \( n \)-step Temporal Difference (TD) with Generalized Advantage Estimation (GAE):
\begin{equation}
\small
V_{\text{target}}(s_t) = \sum_{k=0}^{n-1} \gamma^k r_{t+k} + \gamma^n V(s_{t+n})
\end{equation}

If \( t + n \geq T \), the bootstrap term \( \gamma^n V(s_{t+n}) \) is set to zero to account for episode termination. MSE emphasizes large errors and captures prediction stability, while MAE reflects overall distributional deviation. As the rewards are obtained through simulation, \textbf{these metrics primarily serve as relative indicators of training quality rather than absolute performance benchmarks}.

\begin{figure*}[h]
    \centering
    \includegraphics[width=0.63\linewidth]{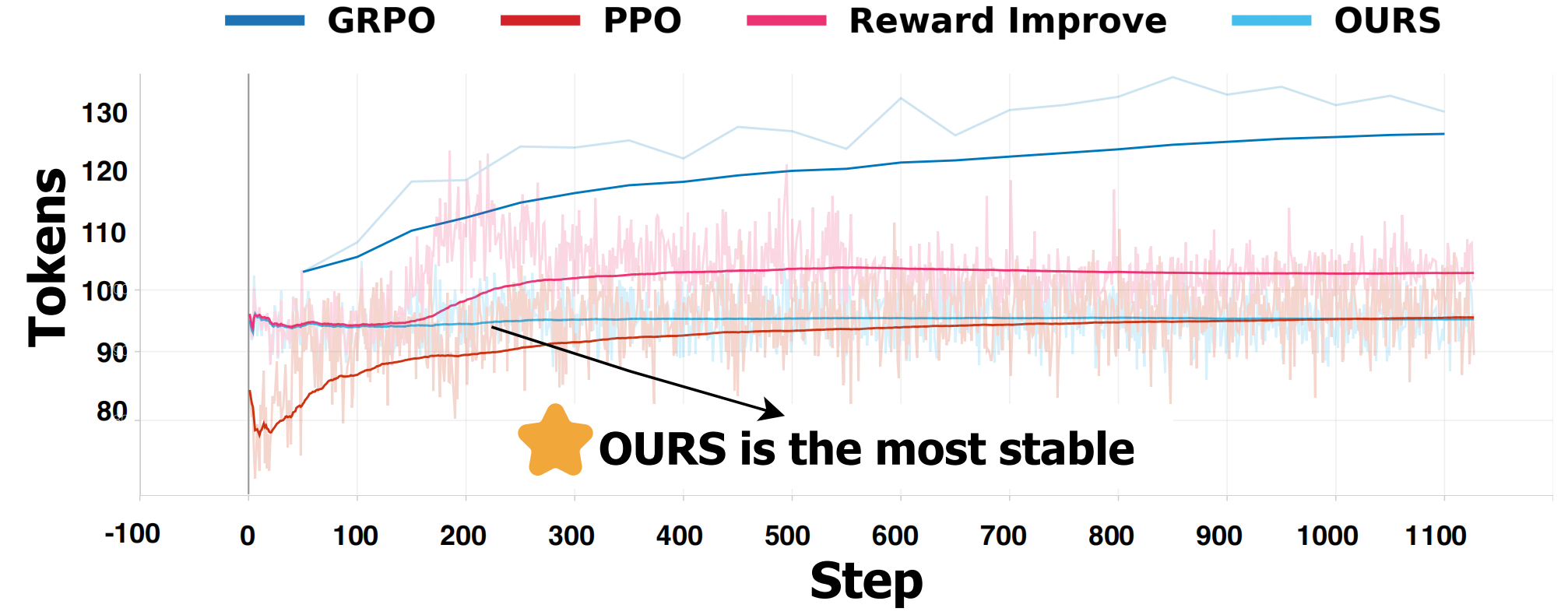}
    \caption{Average response length across training. Our method maintains stable lengths (94 to 95), while PPO and GRPO increase significantly due to length reward hacking.}
    \label{fig:length_curve}
\end{figure*}

\subsubsection{Advantage Visualization } 

To qualitatively assess the robustness introduced by value model training, we visualize token-level advantage values across generated sequences. In reinforcement learning-based language generation, the advantage function quantifies the relative benefit of taking action $a_t$ in state $s_t$, and is computed as:
\begin{equation}
\small
A_t = Q(s_t, a_t) - V(s_t)
\end{equation}
where $Q(s_t, a_t)$ represents the estimated return of taking action $a_t$ in state $s_t$, and $V(s_t)$ denotes the baseline value of the current state. \\
Higher advantage values indicate that the corresponding token exhibits significantly better performance compared to the average policy behavior under the current context. These tokens function as positive learning signals and reflect directions along which the policy should be reinforced. Therefore, visualizing token-level advantages enables us to analyze which parts of the generation are being prioritized during optimization and how the learned policy aligns with high-reward behaviors.

\subsection{Dialogue Task under Model-Based Reward RL Training}
\begin{figure}[h]
    \centering
    \includegraphics[width=1 \linewidth]{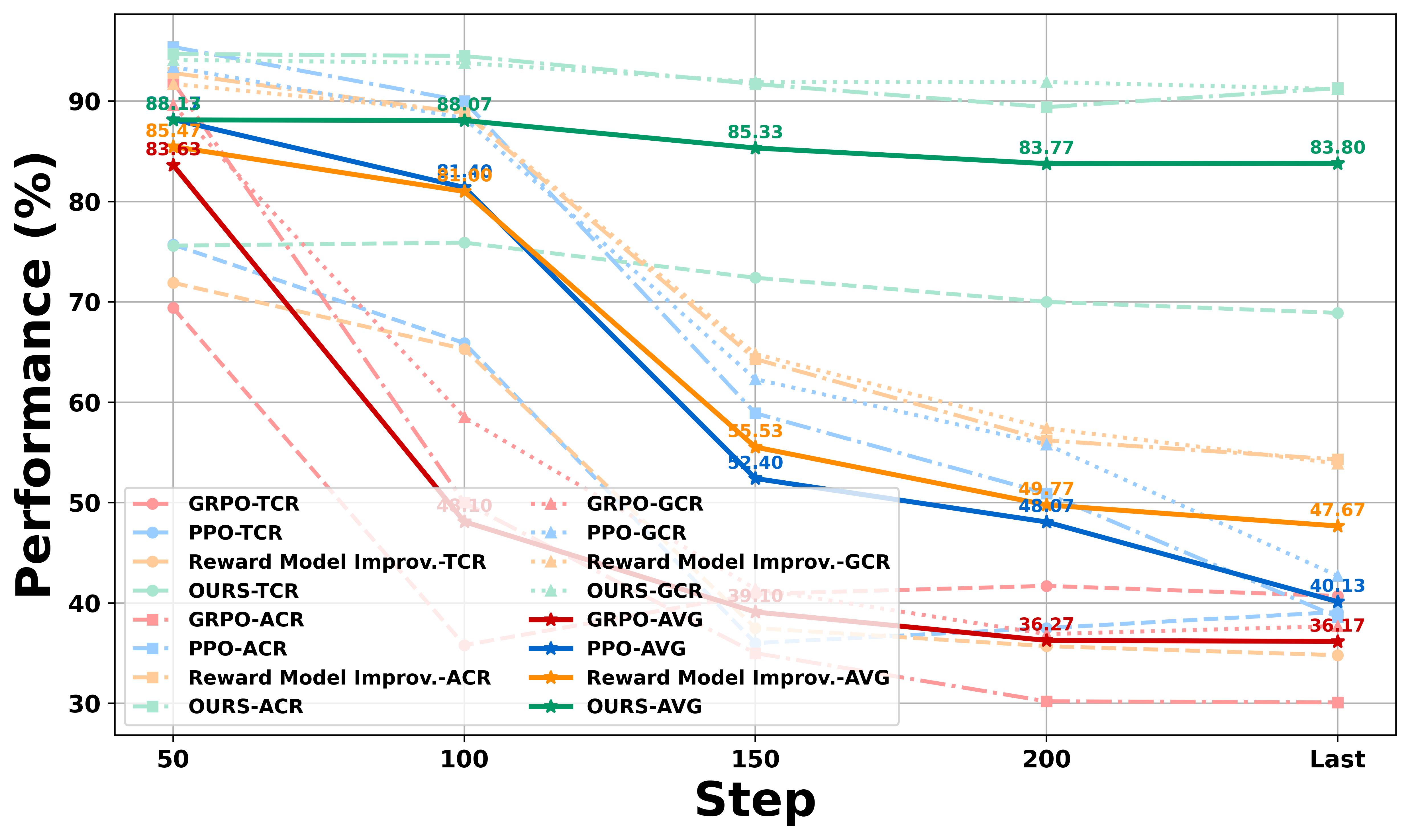}
    \caption{Training performance on dialogue tasks. VRPO (the top lines) improves task completion from 72.1\% to 75.9\%, avoids collapse under noise, and achieves a final average performance of 83.80\%, outperforms GRPO (36.17\%) and PPO (40.13\%) in stability.}
    \label{fig:dialogue_curve}
\end{figure}

To assess the effectiveness of our approach in realistic settings with model-based reward supervision, we conduct multi-turn dialogue experiments on the \textbf{Honor-Dialogue Dataset}. This dataset, constructed by us, contains multi-domain, task-oriented dialogues collected from real-world scenarios. The task requires the model to act as a dialogue assistant, producing natural and effective responses. Under noisy dialogue reward model supervision, the model is trained with reinforcement learning on real multi-turn dialogues. Performance is evaluated through interactions with GPT-4o using the dialogue standards in Section~4.2.1. Dataset statistics are provided in the Appendix, and Figure~\ref{fig:dialogue_curve} shows the training dynamics. All real information in the data has undergone anonymization, and the use of the data has been authorized.

\paragraph{VRPO enhances training stability in real-world general scenarios}
Despite noisy supervision, our method effectively improves dialogue agent performance, surpassing the cold-start baseline from 85.87\% to 88.13\%.  In contrast, PPO and GRPO exhibit severe performance collapse under noisy rewards, with GRPO performing the worst.  This demonstrates our method’s stability and resilience in noisy settings. 
For further experimental data results, please refer to Appendix.

\paragraph{Strong Value Model Helps Reduce Reward Hacking}

A stronger value model helps mitigate reward model hacking by resisting length bias during training. As shown in Figure~\ref{fig:length_curve}, the performance collapse largely stems from length-based reward exploitation. Our method stabilizes advantage estimation, preventing it from scaling with response length. In contrast, PPO and GRPO exhibit sharp length inflation, revealing greater vulnerability to biased rewards.

\paragraph{Value model better corrects model training bias compared to Reward model} We applied our method to the reward model for the experiment. The results show that the value model provides denser supervision, leading to superior correction of the final advantage estimates. Compared to reward model optimization methods, VRPO better preserves training performance in dialogue tasks, achieving an optimal average performance improvement of 75.90\%, with a final average performance exceeding 36.13\%. During training, length fluctuations are more stable, effectively addressing reward biases caused by length during training.

\subsection{Math and Science Task under Rule-Based Rewards RL Training}
\begin{table*}[h]
\centering
\small
\setlength{\tabcolsep}{1.5mm}{
\begin{tabular}{lccccccc|c}
\toprule
\textbf{Domain} & \multicolumn{4}{c}{\textbf{Math}} & \textbf{Factuality} & \textbf{Science} & \textbf{Knowledge} & \textbf{ALL}   \\
\cmidrule(lr){1-1} \cmidrule(lr){2-5} \cmidrule(lr){6-8}
\textbf{Method} & \textbf{MATH500} & \textbf{AIME24} & \textbf{Minerva-Math} & \textbf{AMC23} & 
\textbf{SampleQA} & \textbf{GPQA} & \textbf{HLE} & \textbf{AVG}  \\
\midrule
\multicolumn{9}{l}{\textit{Qwen2.5-7B-Cold Start(Weak Model)}} \\
Base        & 71.60\% & 6.67\%  & 18.75\% & 52.50\% & 2.36\% & 1.45\% & 3.29\% & 22.37\% \\
GRPO        & 50.40\% & 6.67\%  & 13.60\% & 22.50\% & 2.36\% & 2.36\% & 2.97\% & 14.41\% \\
PPO         & 67.40\% & 13.33\% & 19.12\% & 35.83\% & 2.54\% & 2.54\% & 3.43\% & 20.60\% \\
Reinforce++ & 63.80\% & 6.67\%  & 10.66\% & 40.83\% & 2.64\% & 2.17\% & 3.34\% & 18.59\% \\
Dr.GRPO     & \textbf{73.00\%} & 13.33\% & 18.75\% & \textbf{53.33\%} & 2.54\% & 1.27\% & 3.34\% & 23.65\% \\
\rowcolor{gray!30}\textbf{Ours} & 72.20\% & \textbf{23.33\%} & \textbf{19.85\%} & 50.00\% & \textbf{2.82\%} & \textbf{3.44\%} & \textbf{3.94\%} & \textbf{25.08\%} \\
\midrule
\multicolumn{9}{l}{\textit{Qwen3-8B(Strong Model)}} \\
Base        & 87.40\% & 41.67\% & 28.68\% & 75.83\% & 2.89\% & 3.10\% & 2.89\% & 34.64\% \\
GRPO        & 89.20\% & 35.00\% & 28.68\% & 84.17\% & 3.03\% & 2.98\% & 3.24\% & 35.19\% \\
PPO         & 86.40\% & 36.67\% & 30.51\% & 80.00\% & 2.82\% & 2.17\% & 3.29\% & 34.55\% \\
Reinforce++ & 89.00\% & 45.00\% & 27.21\% & 84.17\% & 3.19\% & \textbf{4.35\%} & 3.29\% & 36.60\% \\
Dr.GRPO     & 87.80\% & 45.00\% & 27.94\% & 85.83\% & 2.50\% & 3.99\% & 3.10\% & 36.59\% \\
\rowcolor{gray!30}\textbf{Ours} & \textbf{90.20\%} & \textbf{46.67\%} & \textbf{31.99\%} & \textbf{86.67\%} & \textbf{3.21\%} & \textbf{4.35\%} & \textbf{3.89\%} & \textbf{38.14\%} \\
\bottomrule
\end{tabular}
\caption{Accuracy (\%) on train-time optimization with rule-based rewards across multiple reasoning benchmarks. Our method achieves consistent improvements under both weaker and stronger base models.}
\label{tab:rule_reward_results}
}
\end{table*}

To further assess the effectiveness of our approach in realistic settings where supervision comes from rule-based rewards, we conduct a series of experiments on mathematical and scientific reasoning tasks during training-time optimization. Specifically, we use the Light-R1 dataset \cite{wen2025lightr1curriculumsftdpo} for math tasks and the SuperGPQA\cite{pteam2025supergpqascalingllmevaluation} dataset for scientific tasks. For each dataset, we perform multiple inferences with the original model, followed by voting to filter the results before training. The evaluation is then conducted on math and scientific QA datasets. The test datasets include four math-focused tasks: MATH500 \cite{hendrycks2021measuringmathematicalproblemsolving}, AIME24, Minerva-Math \cite{lewkowycz2022solvingquantitativereasoningproblems}, AMC23, as well as three scientific knowledge tasks: SampleQA \cite{wei2024measuringshortformfactualitylarge}, GPQA , and HLE (Humanity’s Last Exam)\cite{phan2025humanitysexam}. Table~\ref{tab:rule-based-results} reports the accuracy of models before and after RL training, comparing weak and strong models under noisy conditions. Due to the limitation of memory, multiple rounds of generation are carried out here. In each round, the generated length is at most 4096 tokens, and the result of the last round is used for evaluation.

\paragraph{VRPO achieves optimization under noisy supervision across multiple domains} By utilizing a value model with stronger language perception, VRPO consistently outperforms other baseline methods in multiple domains. This improvement is particularly evident in math reasoning and scientific knowledge tasks. For example, after training with the Qwen3-8B model, the average accuracy on math tasks increased from 58.40\% to 60.55\%, and on scientific tasks from 2.96\% to 3.82\%. These results highlight the model’s ability to extract relevant information from noisy or fuzzy feedback and showcase its strong generalization across various domains.

\paragraph{VRPO shows significant effectiveness in scenarios with severe training collapse} In math tasks with a weaker model (Qwen2.5-7B cold-start), model training is heavily impacted by noise, leading to a notable performance drop in GRPO, with accuracy on MATH500 dropping to 50.40\%.  With a stronger model (Qwen3-8B), noise impact decreases, and all methods improve. Our method still provides stable gains, achieving 38.14\% average accuracy on all datasets, outperforming Reinforce++ by 1.54\%. In the weaker model setup, the gain is even higher (6.49\%), demonstrating the advantage of our method under more challenging conditions.
\paragraph{VRPO helps guide the model’s advantage estimates to focus on key words} Figure~\ref{fig:adv-vis} visualizes the advantage estimation for the same response across different methods. While PPO disperses attention across tokens, our model focuses on critical text information. It not only captures the key elements of the response but also shows a more appropriate perception and judgment of its answer.

\subsection{Ablation Experiment and Discussion}

\begin{table*}[h]
\centering
 \small
 \setlength{\tabcolsep}{1mm}{
\begin{tabular}{lcccc|c}
\toprule
\textbf{Method}&\textbf{MATH500}&\textbf{AIME24}&\textbf{Minerva-Math} & \textbf{AMC23}&
\textbf{AVG} \\
\midrule
\multicolumn{6}{l}{\textit{Actor Model}} \\
Cross-Entropy Loss & 73.20\% & 10.00\% & 19.12\% & 50.83\% & 38.29\% \\
CE with Entropy/Perplexity Filtering  & 70.40\% & 13.33\% & 19.12\% & 50.83\% & 38.42\% \\
Entropy + Perplexity Minimization & 69.40\% & 10.00\% & 18.01\% & 54.17\% & 37.90\% \\
\midrule
\multicolumn{6}{l}{\textit{Value Model}} \\
Cross-Entropy Loss & 73.00\% & 10.00\% & 18.38\% & 56.67\% & 39.51\% \\
CE with Entropy/Perplexity Filtering & 72.20\% & 10.00\% & 18.75\% & 57.50\% & 39.61\% \\
\rowcolor{gray!30}\textbf{Entropy + Perplexity Minimization(Ours)} & \textbf{74.40\%} & \textbf{13.33\%} & \textbf{20.22\%} & \textbf{61.67\%} & \textbf{42.41\%} \\
\bottomrule
\end{tabular}
\caption{Comparison of different training objectives on math reasoning tasks under test-time reward perturbation. Entropy- and perplexity-based regularization improves value model robustness (+2.9\%) over standard cross-entropy, confirming its role as a semantic signal regulator.}
\label{tab:language-awareness}
 }
\end{table*}

\paragraph{Enhanced Semantic Awareness rather than Semantic Understanding}

As shown in Table~\ref{tab:language-awareness}, we conduct an ablation study to evaluate how different forms of language-awareness affect model robustness under noisy reward supervision. Our results suggest that directly solving tasks via cross-entropy is suboptimal for the value model. Instead, regularizing with entropy and perplexity, focusing on semantic perception, leads to stronger generalization and robustness.
Specifically, minimizing these losses improves the value model’s ability to identify meaningful signals amid noisy feedback, boosting average accuracy to 42.41\%, compared to 39.51\% under standard cross-entropy training.
This improvement stems from the model’s enhanced alignment with linguistic structure, enabling it to act not as a task solver, but as a semantic signal regulator, assessing responses with higher fidelity and filtering unstable patterns.
While actor improvements help with comprehension, our results show that the value model is more effective in absorbing reward uncertainty, yielding  a 4\% gain over actor-centric training on average.

\paragraph{Partial Token Activation may Strengthen Semantic Learning}
\begin{table}[ht]
\centering
\small
\resizebox{\columnwidth}{!}{
\begin{tabular}{lcccc|c}
\toprule
\makecell{\textbf{Activ-} \\ \textbf{ation}} & 
 \makecell{\textbf{MATH} \\ \textbf{500}} & \makecell{\textbf{AIME} \\ \textbf{24}} & \makecell{\textbf{Minerva} \\ \textbf{Math}} & \makecell{\textbf{AMC} \\ \textbf{23}} & \textbf{AVG} \\
\midrule	
0\%(IB-only) & 72.20\% & 10.00\% & \textbf{20.22\%} & 51.67\% & 38.52\% \\
20\% & 75.00\% & 6.67\% & 18.38\% & 54.17\% & 38.56\% \\
50\% & 73.40\% & 13.33\% & 19.49\% & 55.83\% & 40.51\% \\
\rowcolor{gray!30}\textbf{80\%(Ours)} & \textbf{74.40\%} & 13.33\% & \textbf{20.22\%} & \textbf{61.67\% }& \textbf{42.41\%} \\
100\% & 73.80\% & \textbf{16.67\%} & 19.12\% & 55.00\% & 41.15\% \\
\bottomrule
\end{tabular}%
}
\caption{Impact of partial token activation on performance via entropy/perplexity-based loss, activating 80\% of high-uncertainty tokens yields the best accuracy.}
\label{tab:partialactivation}
\end{table}

We investigate how partially activating high-uncertainty tokens affects robustness in noisy reward training by selectively applying language-aware losses. As shown in Table~\ref{tab:partialactivation}, activating 80\% of high-entropy/perplexity tokens yields the best performance (42.41\%), demonstrating that focusing supervision on uncertain regions strengthens the value model’s alignment with language semantics.
Interestingly, fully activating all tokens (100\%) slightly reduces performance (to 41.15\%), likely due to the inclusion of noisy or uninformative tokens, which destabilize value estimation. This reveals a trade-off: while broader activation promotes semantic sensitivity, indiscriminate supervision can introduce harmful variance. Partial token activation offers a balanced solution that enhances robustness without compromising core value function stability.

\section{Conclusion}
In this study, we rethink the often-overlooked role of the value model in RL frameworks, particularly under noisy reward supervision in LLM post-training. We propose VRPO, a novel training framework that improves the robustness of RL by incorporating information flow filtering and enhancing language perception through guidance from a frozen language model in the value model. This approach effectively corrects reward biases, leading to more accurate advantage estimation. Experiments on math, science and dialogue tasks show that VRPO outperforms baseline methods such as PPO and GRPO, highlighting the untapped potential of the value model in robust RL training.

\section*{Limitations}
In this section, we discuss the potential threats to the validity of our method. VRPO does not solve all forms of reward noise, but offers a principled way to make the value model more robust and semantically aware in the face of noisy supervision. And due to limited computational resources, the experiments with rule-based rewards were only conducted on mathematics and science tasks, where each model was trained and evaluated on a single dataset. Moreover, the ablation studies were primarily performed during test-time optimization under rule-based reward settings. To mitigate this potential threat, we further conducted experiments across a wide range of datasets and tasks, and performed ablation studies under rule-based reward conditions during training-time optimization as well. Although current results indicate no significant differences between the two settings, this broader evaluation helps ensure the robustness of our conclusions.

\section*{Acknowledgments}
The authors wish to thank the anonymous reviewers for their helpful comments. This work was partially funded by National Natural Science Foundation of China (No.62521004, 62476061, 62576106, 62376061).


\bibliography{custom}

\appendix

\section{Additional Details for VRPO}
\label{sec:appendix}

\begin{algorithm}
\caption{VRPO training process}
\label{alg:ppo-ib}
\begin{algorithmic}
\REQUIRE Dataset $\mathcal{D} = \{(x_t, a_t, r_t)\}_{t=1}^T$, policy $\pi_\theta$, decoder model $q_\psi$, encoder model $f_\phi$, bottleneck prior $r(z)$; coefficients $\beta, \lambda_{\text{ent}}, \lambda_{\text{ppl}}$
\ENSURE Actor model $\pi_\theta$ , value model $f_\phi, q_\psi$
\FOR{each VRPO iteration}
    \STATE Sample trajectories using current policy $\pi_\theta$
    \STATE Compute rewards $\{r_t\}$ and actions $\{a_t\}$
    \STATE Compute policy ratio: $\rho_t = \frac{\pi_\theta(a_t|x_t)}{\pi_{\theta_{\text{old}}}(a_t|x_t)}$
    \FOR{each timestep $t$}
        \STATE Encode latent: sample $\epsilon \sim \mathcal{N}(0, I)$, compute $z_t = f^\mu_\phi(x_t) + f^\sigma_\phi(x_t) \cdot \epsilon$
        \STATE Predict value: $\hat{V}_t = q_\psi(z_t)$
    \ENDFOR
    \STATE Compute GAE advantage: $A_t = \text{GAE}(r_t, \hat{V}_t, \hat{V}_{t+1})$
    \STATE Compute returns: $R_t = A_t + \hat{V}_t$

    \STATE
    \STATE \textbf{Value Model Update}
    \STATE
    
    \STATE \textit{// Value Loss with IB Structure}
    \STATE Compute value loss: $\mathcal{L}_{\text{MSE}} = \frac{1}{T} \sum_t (q_\psi(f\phi(x_t)) - R_t)^2$
    \STATE Compute KL loss: $\mathcal{L}_{\text{KL}} = \frac{1}{T} \sum_t \text{KL}(p_\phi(z_t|x_t) \| r(z))$
    \STATE \textit{// Semantic Alignment via Frozen LM Head}
    \FOR{each timestep $t$}
        \STATE Get token distribution $P_V(y_t | x_t)$ from frozen LM head
        \STATE Identify high-uncertainty tokens:
        \STATE \quad $T_{\text{ent}} = \{t : H[P_V(y_t | x_t)] > \hat{T}_{\text{entropy}}\}$
        \STATE \quad $T_{\text{ppl}} = \{t : -\log P_V(y_t = y_t^* | x_t) > \hat{T}_{\text{perplexity}} \}$
        \STATE Compute entropy loss: $L_{\text{ent}} = \sum_{t \in T_{\text{ent}}} H[P_V(y_t | x_t)]$
        \STATE Compute perplexity loss: $L_{\text{ppl}} = \sum_{t \in T_{\text{ppl}}} -\log P_V(y_t = y_t^* | x_t)$
    \ENDFOR
    \STATE Compute semantic loss: $\mathcal{L}_{\text{sem}} = \lambda_{\text{ent}} \cdot L_{\text{ent}} + \lambda_{\text{ppl}} \cdot L_{\text{ppl}}$
    \STATE Combine value loss: 
        $\mathcal{L}_{\text{value}} = \mathcal{L}_{\text{MSE}} + \beta \cdot \mathcal{L}_{\text{KL}} + \mathcal{L}_{\text{sem}}$
    \STATE Update value model $\{\phi, \psi\}$ using $\nabla \mathcal{L}_{\text{value}}$

    \STATE
    \STATE \textbf{Actor Model Update}
    \STATE
    
    \STATE Compute clipped PPO objective: 
        $\mathcal{L}_{\text{PPO}} = \mathbf{E}_t \left[\min\left( \rho_t A_t, \text{clip}(\rho_t, 1 - \epsilon, 1 + \epsilon) A_t \right)\right]$
    \STATE Update actor model $\pi_\theta$, using $\nabla \mathcal{L}_{\text{PPO}}$
\ENDFOR
\end{algorithmic}
\end{algorithm}

\subsection{Pseudocode}
The full algorithm of VRPO is detailed in Algorithm ~\ref{alg:ppo-ib}.
\section{Additional Experimental Details}
\label{appendix:setup}
\subsection{Setup}
\subsubsection{Dataset Construction}

\begin{table*}[t]
\centering
\small
\resizebox{2\columnwidth}{!}{
\begin{tabular}{lccccccccccc|c}
\toprule
\textbf{Scenario} & \textbf{Wealth} & \textbf{Rental} & \textbf{Insurance} & \textbf{Food} & \textbf{Express} & \textbf{Promotion} & \textbf{Loan} & \textbf{Housing} & \textbf{Service} & \textbf{Product} & \textbf{General} & \textbf{Avg} \\
\midrule
Dialogue Count & 87 & 99 & 138 & 120 & 215 & 66 & 70 & 87 & 67 & 69 & 92 & 94.73 \\
Avg Turns      & 5.40 & 4.22 & 5.22 & 3.37 & 3.76 & 4.55 & 4.33 & 4.68 & 5.46 & 4.44 & 3.46 & 4.44 \\
\bottomrule
\end{tabular}
}
\caption{
Dataset statistics across 11 real-world service scenarios.
The validation set covers diverse interaction types with varying dialogue lengths, enabling reliable evaluation of robustness and generalization under realistic settings.
}
\label{tab:validation-data-statistics}
\end{table*}

\begin{table*}[ht]
\centering
\small
\resizebox{0.7\linewidth}{!}{
\begin{tabular}{lcccccc|cc}
\toprule
\textbf{Category} & \multicolumn{4}{c}{\textbf{Human Annotators}} & \multicolumn{2}{c|}{\textbf{GPT Models}} & \textbf{Human AVG} & \textbf{Model AVG}  \\
\cmidrule(lr){1-1} \cmidrule(lr){2-5} \cmidrule(lr){6-7} \cmidrule(lr){8-9}
Score & 4.595  & 4.689  & 4.490  & 4.330  & 4.510  & 4.501  & 4.526     & 4.505     \\
\bottomrule
\end{tabular}
}
\caption{Comparison of goal completion rate scores between human annotators and GPT-4o. 
Human annotators achieve an average score of 4.526, while GPT-4o models reach 4.505, showing a negligible performance gap.}
\label{tab:human-model-agreement}
\end{table*}
\paragraph{Rule-based Setting.}
For \textbf{test-time augmentation}, we use four math benchmarks: AIME 2024, AMC 2023, Minerva-Math, and MATH-500, as well as three QA datasets: GPQA-Extended, SimpleQA, and Humanity’s Last Exam (HLE).

For \textbf{training-time augmentation}, as an illustrative example based on the Qwen3-8B model, pseudo-labels for mathematical tasks are generated from 39,000 samples in the Light-R1 \cite{wen2025lightr1curriculumsftdpo} dataset. After 5 rounds of majority voting, samples with at least 3 identical responses are retained, yielding 31,209 instances for reinforcement learning training.  For scientific tasks, pseudo-labels are generated from 26,529 samples in the SuperGPQA dataset. After the same 5-round majority voting procedure, 10,075 samples with at least 2 consistent responses are preserved for RL training.  
Evaluation is conducted on non-overlapping mathematical benchmarks to ensure fairness.  

\paragraph{Model-based Setting.}
A real-world dialogue dataset (Honor-Dialogue) is used. The reward model is trained on 36,000 labeled samples, with 3,000 for validation. The policy model is fine-tuned and trained on 80,000 additional conversations, with 8,000 for evaluation.This dataset contains various dialogue tasks in real-life scenarios. It truly reflects the actual conditions of real situations. Such multi-category real-scenario dialogue task data is not available in other datasets. All real information in the data has undergone anonymization, and the use of the data has been authorized and reviewed by the ethics committee.

Honor-Dialogue dataset is constructed based on a goal-driven scenario-oriented design paradigm. For the data within each domain, we construct realistic caller inputs that include scenario-matched latest messages and conversation history, as well as corresponding standardized outputs that comply with the requirements of the target goal.   We explicitly mark the dialogue state, response content, and matched target to ensure the quality of the supervised training data.
A representative example of the Honor-Dialogue dataset is presented below.(Figure ~\ref{fig:dialogue-example})    The notation xxx denotes the masking of sensitive information such as user ID and contact details, which is implemented to comply with data privacy regulations.

\subsubsection{Baseline Initialization}

\paragraph{Rule-based Setting.}
\textbf{Test-time augmentation:} Qwen2.5-7B-Base (math task) and Qwen2.5-7B-Instruct (QA task), fine-tuned on OpenR1-Math and S1.1\cite{muennighoff2025s1simpletesttimescaling}, respectively.
\textbf{Training-time augmentation:} Qwen2.5-7B-Cold Start, Qwen3-1.7B , Qwen3-8B and  LLaMA 3.1-8B-Instruct are used as baselines.The Qwen2.5-7B-Cold Start model has the same model settings as those in the Test-time augmentation.

\paragraph{Model-based Setting.}
Both the policy and reward models are initialized from Qwen3-8B, fine-tuned on the Honor-Dialogue dataset.

\subsubsection{Training Configuration}
\paragraph{Training parameters.}In the majority voting during testing, 32 samples are used, while 5 samples are used for voting during training. The weights for the auxiliary losses (entropy and perplexity) are set to 0.5, making 80\% of the labels effective. The RL training runs for 1 iteration and the RL setup adopts a training batch size of 128 and a rollout batch size of 128, with actor and critic learning rates of $5\times10^{-7}$ and $5\times10^{-6}$. During both training and testing, the model generates dialogue tasks of length 4096 for reasoning tasks and 1024 for dialogue tasks. The experiments are conducted on 8 * NVIDIA A100 80GB GPUs.

\subsubsection{Dialogue Task Evaluation}

Our dialogue evaluation leverages GPT-4o, but the model is not utilized as a free-form judge. Instead, it implements a rigorous rubric-based evaluation protocol designed to mitigate subjectivity and enhance reproducibility. This rubric explicitly defines evaluation metrics, adopts a 1–5 scoring scale with clear criteria, includes two scoring examples and a standardized output format, and the prompt incorporates a well-structured process along with comprehensive dialogue content and contextual information. GPT-4o functions solely as an automated evaluator applying this fixed rubric, rather than an unconstrained scorer.

To further ensure reliability, we conducted cross-validation through two key steps: \textbf{1)} multiple sampling runs to verify the stability of rubric execution, and \textbf{2)} spot-checked human evaluations that demonstrated high agreement with rubric-based scores. Specifically, for the comparative experiment between human annotators and the model, we validated the protocol using 1,110 data samples covering 10 scenarios. The statistical details of the validation dataset are presented in Table \ref{tab:validation-data-statistics}.
Four independent annotators scored these samples strictly in accordance with the rubric. They are professional data annotators in the company. Table \ref{tab:human-model-agreement} presents the experimental results for goal completion rate scoring (full score: 5) in dialogue tasks.
As shown in Table \ref{tab:human-model-agreement}, the mean score assigned by human annotators is extremely close to that of the model, with a difference of only 0.021 points. These findings confirm strong consistency between human and model scores, validating the effectiveness of our rubric. We acknowledge that fully establishing external validity necessitates additional independent annotators. We are in the process of releasing our rubric, evaluation prompts, and real-dialogue dataset to facilitate replication.
To illustrate the evaluation logic for dialogue performance, the core prompt section regarding the assessment of conversation logicality is provided as follows in Figure \ref{fig:scoring-guidelines}, which is a part of goal completion rate.

\subsection{Additional experimental details in the Dialogue Task}

To further analyze the behavior of different methods in the dialogue task, Table~\ref{tab:dialogue-results} presents the full training trajectory across various steps. A notable observation is that both PPO and GRPO experience severe performance degradation during training, likely due to reward over-optimization or instability, resulting in significantly lower final scores than both the initial model and our method. Specifically, their final average scores drop to 40.13\% and 36.17\%, respectively.

In contrast, our proposed method (VRPO) maintains high robustness throughout training. While the final performance slightly decreases compared to the cold-start model (from 85.87\% to 83.80\%), our method avoids collapse and achieves high task performance in early training stages. For example, at step 50 and 100, VRPO reaches 75.90\% task completion rate (TCR) and an overall average (AVG) score of 88.13\%, outperforming all baselines at these stages.

These results suggest that VRPO effectively preserves model capabilities under heavy noisy supervision and raises the lower bound of post-training performance. This stabilizing effect makes it significantly less prone to reward hacking or collapse compared to standard PPO approaches.

\begin{table}[h]
\centering
\small
\begin{tabular}{lcccc}
\toprule
\textbf{Step} & \textbf{TCR} & \textbf{ACR} & \textbf{GCR} & \textbf{AVG} \\
\midrule
\multicolumn{5}{l}{\textit{Cold Start}} \\
0   & 72.10\% & 94.30\% & 91.20\% & 85.87\% \\
\midrule
\multicolumn{5}{l}{\textit{GRPO}} \\
 50   & 69.40\% & 91.90\% & 89.60\% & 83.63\% \\
 100  & 35.80\% & 50.00\% & 58.50\% & 48.10\% \\
 150  & 40.90\% & 35.00\% & 41.40\% & 39.10\% \\
 200  & 41.70\% & 30.20\% & 36.90\% & 36.27\% \\
\rowcolor{gray!30}Last & 40.70\% & 30.10\% & 37.70\% & 36.17\% \\
\midrule
\multicolumn{5}{l}{\textit{PPO}} \\
 50   & 75.70\% & \textbf{95.40\%} & 93.40\% & \textbf{88.17\%} \\
 100  & 65.90\% & 90.00\% & 88.30\% & 81.40\% \\
 150  & 36.00\% & 58.90\% & 62.30\% & 52.40\% \\
 200  & 37.50\% & 50.90\% & 55.80\% & 48.07\% \\
\rowcolor{gray!30}Last & 39.10\% & 38.60\% & 42.70\% & 40.13\% \\
\midrule
\multicolumn{5}{l}{\textit{Reward model improvement method base PPO}} \\
 50   & 71.90\% & 92.80\% & 91.70\% & 85.47\% \\
 100  & 65.30\% & 88.80\% & 88.90\% & 81.00\% \\
 150  & 37.50\% & 64.30\% & 64.80\% & 55.53\% \\
 200  & 35.70\% & 56.20\% & 57.40\% & 49.77\% \\
\rowcolor{gray!30}Last & 34.80\% & 54.30\% & 53.90\% & 47.67\% \\
\midrule
\multicolumn{5}{l}{\textbf{\textit{OURS}}} \\
 50   & 75.60\% & 94.70\% & \textbf{94.10\%} & 88.13\% \\
 100  & \textbf{75.90\%} & 94.50\% & 93.80\% & 88.07\% \\
 150  & 72.40\% & 91.70\% & 91.90\% & 85.33\% \\
 200  & 70.00\% & 89.40\% & 91.90\% & 83.77\% \\
\rowcolor{gray!30}Last & 68.90\% & 91.30\% & 91.20\% & 83.80\% \\
\bottomrule
\end{tabular}
\caption{Trajectory-level generalization performance across different methods and training steps. Our method maintains high performance even in last stages, beyond GRPO, PPO, and the reward model improvement method based on PPO.}
\label{tab:dialogue-results}
\end{table}

\subsection{Additional experimental details in the Math and Science Task}

\paragraph{Comparison with Recent Robust Advantage-Based RL Methods}

\begin{table*}[h]
\centering
\small
\setlength{\tabcolsep}{1.5mm}{
\begin{tabular}{lccccccc|c}
\toprule
\textbf{Domain} & \multicolumn{4}{c}{\textbf{Math}} & \textbf{Factuality} & \textbf{Science} & \textbf{Knowledge} & \textbf{ALL}   \\
\cmidrule(lr){1-1} \cmidrule(lr){2-5} \cmidrule(lr){6-8}
\textbf{Method} & \textbf{MATH500} & \textbf{AIME24} & \textbf{Minerva-Math} & \textbf{AMC23} & 
\textbf{SampleQA} & \textbf{GPQA} & \textbf{HLE} & \textbf{AVG}  \\
\midrule
\multicolumn{9}{l}{\textit{Qwen2.5-7B-Cold Start(Weak Model)}} \\
Base   & 71.60\% & 6.67\% & 18.75\%      & 52.50\% & 2.36\% & 1.45\%    & 3.29\% & 22.37\% \\
KTAE   & 68.60\% & 6.67\% & 19.12\%      & 48.33\% & 1.73\% & 2.36\%    & 3.24\% & 21.44\% \\
$\lambda$-GRPO & \textbf{74.00\%} & 10.00\% & 18.38\% & \textbf{56.67\%} & \textbf{2.94\%} & 3.26\% & 3.34\% & 24.08\% \\
\rowcolor{gray!30}\textbf{Ours} & 72.20\% & \textbf{23.33\%} & \textbf{19.85\%} & 50.00\% & 2.82\% & \textbf{3.44\%} & \textbf{3.94\%} & \textbf{25.08\%} \\
\midrule
\multicolumn{9}{l}{\textit{Qwen3-8B(Strong Model)}} \\
Base   & 87.40\% & 41.67\% & 28.68\%      & 75.83\% & 2.89\% & 3.10\%    & 2.89\% & 34.64\% \\
KTAE   & 86.00\% & 38.33\% & 29.41\%      & 78.33\% & 3.17\% & 3.99\%    & 2.64\% & 34.55\% \\
$\lambda$-GRPO & 90.00\% & \textbf{50.00\%} & 31.62\% & 80.00\% & 2.77\% & \textbf{4.35\%} & 3.57\% & 37.47\% \\
\rowcolor{gray!30}\textbf{Ours} & \textbf{90.20\%} & 46.67\% & \textbf{31.99\%} & \textbf{86.67\%} & \textbf{3.21\%} & \textbf{4.35\%} & \textbf{3.89\%} & \textbf{38.14\%} \\
\bottomrule
\end{tabular}
\caption{
Comparative accuracy (\%) of train-time optimization across mathematical and scientific reasoning benchmarks with recent advantage-based robust RL methods.
}
\label{tab:rule_reward_resultsadv}
}
\end{table*}

We further compare VRPO with recent and conceptually related robust RL methods that explicitly target advantage shaping or robustness, including KTAE and $\lambda$-GRPO. While all these approaches aim to mitigate noisy or biased advantages, they differ fundamentally in how token-level credit assignment is addressed.

KTAE adopts an explicit keyword-oriented strategy, directly identifying salient tokens and reweighting advantages based on rule-based key-token selection. In contrast, VRPO does not rely on predefined or externally extracted keywords. Instead, it leverages semantic guidance through entropy- and perplexity-based losses applied to the value model, encouraging advantage estimation to implicitly attend to semantically informative tokens via representation learning. This enables VRPO to capture task-relevant signals in a softer, model-driven manner rather than through explicit token masking or selection.

Similarly, $\lambda$-GRPO introduces robustness by explicitly penalizing response length through a length-dependent weighting in the softmax computation, thereby discouraging degenerate long responses. VRPO complements this approach by strengthening the value model itself: through the information bottleneck and semantic regularization, the value model learns to better perceive and encode the semantic structure of generated text, allowing advantage estimates to adaptively reflect the quality and relevance of content rather than relying on a length-based prior.

Empirically, these conceptual differences translate into consistent performance gains. Under the same train-time optimization setting described in Section~4.1, VRPO achieves the highest average accuracy across both Qwen-2.5-7B Cold-Start and Qwen-3-8B models (Table~\ref{tab:rule_reward_resultsadv}). In the Qwen-2.5-7B setting, VRPO improves the overall average accuracy to 25.08\%, outperforming KTAE (21.44\%) and $\lambda$-GRPO (24.08\%), with gains observed across both mathematical and scientific benchmarks. Similar trends persist for the stronger Qwen-3-8B model, where VRPO again attains the best overall performance (38.14\%), surpassing both KTAE and $\lambda$-GRPO.

Taken together, these results suggest that VRPO offers a complementary robustness mechanism: rather than explicitly constraining advantages via explicit token rules or length penalties, it enhances the semantic sensitivity and noise-filtering capacity of the value model itself, leading to more reliable advantage estimation under noisy supervision.

\paragraph{VRPO consistently improves reasoning performance under rule-based noisy rewards} To further evaluate the effectiveness of our approach in rule-based reward scenarios, we conducted a series of experiments covering both mathematical and factual reasoning domains in test-time inference settings. Specifically, we trained and evaluated the model using seven datasets. For each dataset, we performed multiple inferences with the original model and conducted voting-based result filtering, then trained on the filtered data before evaluating the model on the respective datasets. The datasets include four math-focused tasks: MATH500 \cite{hendrycks2021measuringmathematicalproblemsolving}, AIME24, Minerva-Math \cite{lewkowycz2022solvingquantitativereasoningproblems}, AMC23, and three factual knowledge tasks: SampleQA \cite{wei2024measuringshortformfactualitylarge}, GPQA, and HLE (Humanity's Last Exam). Table~\ref{tab:rule-based-results} presents accuracy after reasoning-enhanced training using different RL methods.
\begin{figure}[h]
    \centering
    \includegraphics[width= \linewidth]{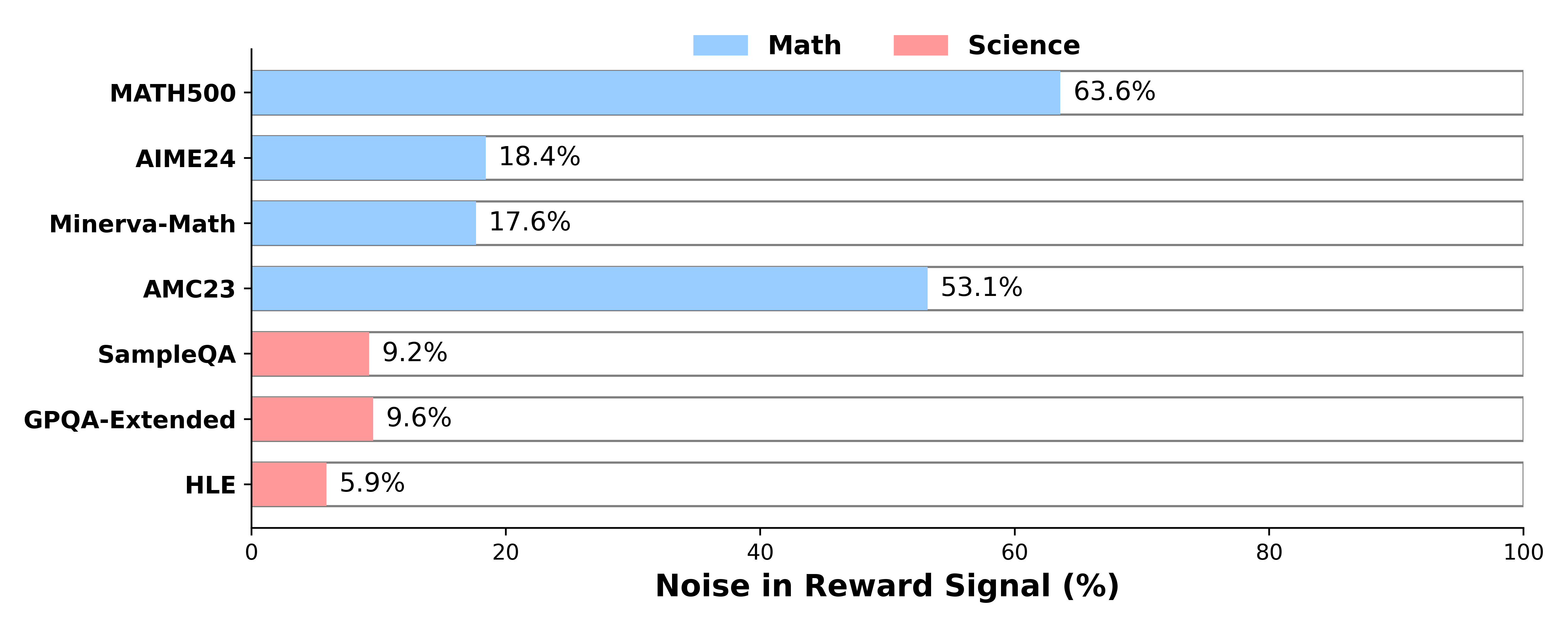}
    \caption{Noise statistics in the math and science tasks with rule-based reward supervision during test-time optimization. }
    \label{fig:dialogue_curve}
\end{figure}

\begin{figure*}[h]
\small
\centering
\includegraphics[width=0.9\linewidth]{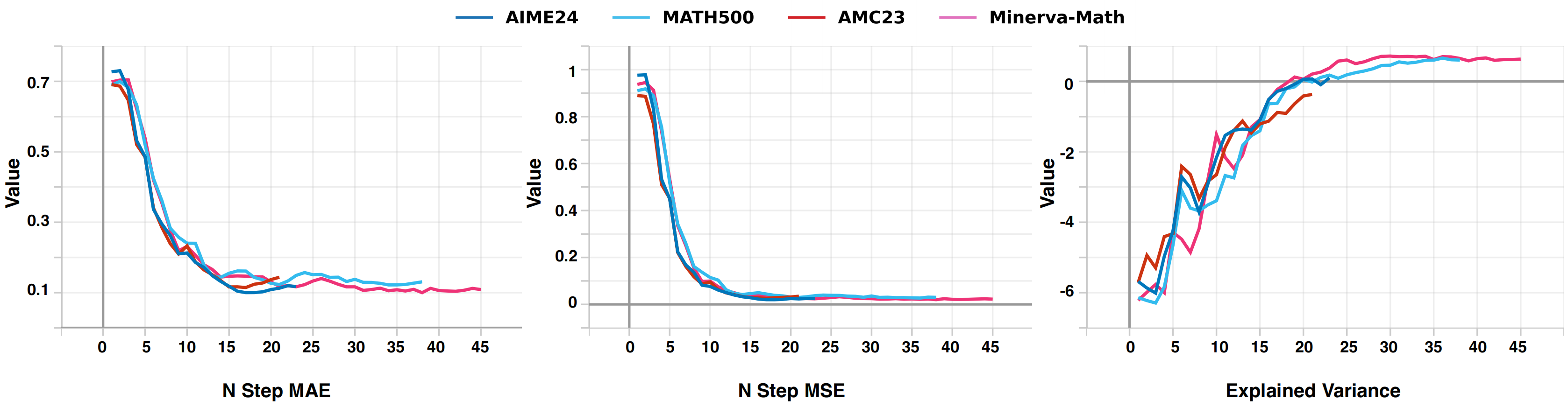}
\caption{Prediction error of the value model across training steps}
\label{fig:error-curve}
\end{figure*}

\begin{table*}[h]
 \small
\centering
 \setlength{\tabcolsep}{1mm}{
\begin{tabular}{lccccccc|c}
\toprule
\textbf{Domain} & \multicolumn{4}{c}{\textbf{Math}} & \textbf{Factuality} & \textbf{Science} & \textbf{Knowledge}& \textbf{ALL}    \\
\cmidrule(lr){1-1} \cmidrule(lr){2-5} \cmidrule(lr){6-8}
\textbf{Method} & \textbf{MATH500} & \textbf{AIME24} & \textbf{Minerva-Math} & \textbf{AMC23} & 
\textbf{SampleQA} & \textbf{GPQA} & \textbf{HLE} & \textbf{AVG} \\
\midrule
Cold Start & 71.60\% & 6.67\% & 18.75\% & 52.50\% & 1.62\% & 2.53\% & 2.55\% & 22.32\% \\
GRPO       & 73.40\% & 11.67\% & 19.49\% & 56.67\% & 2.47\% & 3.08\% & 3.80\% & 24.37\% \\
PPO        & 73.60\% & 10.00\% & 19.49\% & 55.00\% & 2.43\% & 2.54\% & 3.56\% & 23.80\% \\
\rowcolor{gray!30}\textbf{Ours}       & \textbf{74.40\%} & \textbf{13.33\%} & \textbf{20.22\%} & \textbf{61.67\%} & \textbf{2.61\%} & \textbf{4.17\%} & \textbf{4.45\%} & \textbf{25.84\%} \\
\bottomrule
\end{tabular}
\caption{Accuracy (\%) on test-time optimization with rule-based rewards. Our method achieves gains across various datasets and domains.}
\label{tab:rule-based-results}
 }
\end{table*}
Despite noisy feedback, our value model learns effectively. As shown in Figure~\ref{fig:error-curve}, the explained variance of our method increases steadily during training, indicating closer alignment between predicted and GAE-derived returns. This trend is accompanied by a consistent drop in reward prediction error and reduction in high-variance outliers.
As shown in Table~\ref{tab:rule-based-results}, our method consistently outperforms other baseline methods in various domains by leveraging a strong value model with semantic-aware filtering. This improvement is evident in both mathematical and knowledge reasoning tasks. For instance, accuracy on AIME24 increases from 6.67\% to 13.33\%, and on HLE from 2.55\% to 4.45\%. These results highlight the model’s ability to extract relevant information from noisy or ambiguous feedback and demonstrate its robust generalization across diverse domains.

\subsection{Additional Ablation Experiment}

\begin{table}[ht]
\centering
 \small
\setlength{\tabcolsep}{1mm}{
\begin{tabular}{lcccc|c}
\toprule
\textbf{Method} &
\makecell{\textbf{MATH} \\ \textbf{500}} & 
\makecell{\textbf{AIME} \\ \textbf{24}} & 
\makecell{\textbf{Minerva} \\ \textbf{Math}} & 
\makecell{\textbf{AMC} \\ \textbf{23}} & 
\textbf{AVG} \\
\midrule
Base & \textbf{49.80\%} & \textbf{10.00\%} & 17.28\% & 24.17\% & 25.31\% \\
GRPO & 49.20\% & 6.67\% & 16.18\% & \textbf{27.50\%} & 24.89\% \\
PPO & 46.20\% & 6.67\% & \textbf{18.01\%} & 25.00\% & 23.97\% \\
\rowcolor{gray!30}\textbf{Ours} & 49.40\% & \textbf{10.00\%} & 17.65\% & \textbf{27.50\%} & \textbf{26.14\%} \\
\bottomrule
\end{tabular}
\caption{Test-time optimization results on mathematical reasoning benchmarks using LLaMA 3.1-8B-Instruct under noisy rule-based reward supervision. Our method is the only one that improves upon the base model, demonstrating superior robustness under noisy feedback.}
\label{tab:testtime}
}
\end{table}

\begin{table}[ht]
\centering
\small
\setlength{\tabcolsep}{1mm}{
\begin{tabular}{lcccc|c}
\toprule
\textbf{Method} &
\makecell{\textbf{MATH} \\ \textbf{500}} & 
\makecell{\textbf{AIME} \\ \textbf{24}} & 
\makecell{\textbf{Minerva} \\ \textbf{Math}} & 
\makecell{\textbf{AMC} \\ \textbf{23}} & 
\textbf{AVG} \\
\midrule
Base & \textbf{49.80\%} & \textbf{10.00\%} & 17.28\% & 24.17\% & \textbf{25.31\%} \\
GRPO & 42.40\% & 3.33\% & 17.28\% & \textbf{32.50\%} & 23.88\% \\
PPO & 41.00\% & 6.67\% & 17.65\% & 25.83\% & 22.79\% \\
\rowcolor{gray!30}\textbf{Ours} & 41.40\% & \textbf{10.00\%} & \textbf{18.01\%} & 30.00\% & 24.85\% \\
\bottomrule
\end{tabular}
\caption{Training-time optimization results on the math task using LLaMA 3.1-8B-Instruct under noisy reward conditions. Due to weak base performance and high noise, all methods show a performance drop, but ours degrades the least and even improves results on Minerva-Math and AMC23.}
\label{tab:traintime}
}
\end{table}
\paragraph{VRPO effectively mitigates noise in mathematical tasks when applied to the LLaMA model series} Recent studies\cite{wu2025reasoningmemorizationunreliableresults} suggest that Qwen models may be less reliable for mathematical reasoning. Therefore, we design experiments using the LLaMA 3.1-8B-Instruct model.  
Tables~\ref{tab:testtime} and~\ref{tab:traintime} present its performance on four mathematical reasoning benchmarks under rule-based reward supervision. Specifically, Table~\ref{tab:testtime} reports results in the test-time optimization setting, while Table~\ref{tab:traintime} shows training-time optimization outcomes.  

In the test-time optimization setting, only our proposed method achieves improvements over the base model, whereas PPO and GRPO both suffer performance degradation, underscoring VRPO’s robustness under noisy supervision.  
In the training-time optimization setting, due to limited model capacity and the prevalence of noisy samples (with overall data accuracy of only 42.33\%), all methods experience performance drops. Nevertheless, our method exhibits the smallest decline and even achieves gains on Minerva-Math and AMC23, demonstrating stronger resilience in noisy training environments.  
These findings further highlight the generalizability of VRPO across different model architectures, confirming its effectiveness beyond Qwen models.

\paragraph{Effect of Variational Bottleneck on Noise Filtering}

\begin{table*}[htbp]
\centering
\small
\setlength{\tabcolsep}{1mm}{
\begin{tabular}{lccccccc}
\toprule
\textbf{Domain} & \multicolumn{4}{c}{\textbf{Math}} & \textbf{Factuality} & \textbf{Science} & \textbf{Knowledge}   \\
\cmidrule(lr){1-1} \cmidrule(lr){2-5} \cmidrule(lr){6-8}
\textbf{Method} & \textbf{MATH500} & \textbf{AIME24} & \textbf{Minerva-Math} & \textbf{AMC23} & 
\textbf{SampleQA} & \textbf{GPQA} & \textbf{HLE}  \\
\midrule
Zero bottleneck(semantic-only) & 70.20\% & \textbf{16.67\%} & 19.49\% & 50.83\% & 2.47\% & 2.17\% & 3.80\%  \\
Two bottleneck  & 73.20\% & \textbf{16.67\%} & \textbf{21.32\%} & 60.00\% & 2.45\% & 3.44\% & 3.62\%  \\
\rowcolor{gray!30}One bottleneck(\textbf{Ours}) & \textbf{74.40\%} & 13.33\% & 20.22\% & \textbf{61.67\%} & \textbf{2.61\%} & \textbf{4.17\%} & \textbf{4.45\%} \\
\bottomrule
\end{tabular}
\caption{Performance comparison across domains and datasets using different bottleneck strategies.Semantic-only indicates that the experiment uses only the configuration with semantic loss improvements.}
\label{tab:bottleneck}
}
\end{table*}
\begin{table*}[h]
\centering
\small
\setlength{\tabcolsep}{1.5mm}{
\begin{tabular}{lccccccc|c}
\toprule
\textbf{Domain} & \multicolumn{4}{c}{\textbf{Math}} & \textbf{Factuality} & \textbf{Science} & \textbf{Knowledge} & \textbf{ALL}   \\
\cmidrule(lr){1-1} \cmidrule(lr){2-5} \cmidrule(lr){6-8}
\textbf{Method} & \textbf{MATH500} & \textbf{AIME24} & \textbf{Minerva-Math} & \textbf{AMC23} & 
\textbf{SampleQA} & \textbf{GPQA} & \textbf{HLE} & \textbf{AVG}  \\

\midrule
\multicolumn{9}{l}{\textit{Qwen3-1.7B}} \\
Base        & 84.80\% & 35.00\% & 19.49\% & 69.17\% & 1.78\% & 2.36\% & 2.87\% & 30.78\% \\
GRPO        & 84.20\% & 33.33\% & 22.79\% & \textbf{79.17\%} & 1.60\% & 1.99\% & 3.94\% & 32.43\% \\
PPO         & 83.80\% & 35.00\% & 22.06\% & 78.33\% & \textbf{1.91\%} & 2.54\% & 3.94\% & 32.51\% \\
Reinforce++ & 84.40\% & 41.67\% & 23.53\% & 75.83\% & 1.48\% & 2.36\% & 3.52\% & 33.26\% \\
Dr.GRPO     & 83.40\% & 36.67\% & 22.43\% & 75.83\% & 1.50\% & 2.36\% & 3.10\% & 32.19\% \\
KTAE   & 64.40\% & 18.33\% & 18.01\%      & 56.67\% & 1.64\% & \textbf{2.90\%}    & \textbf{4.36\%} & 23.76\% \\
$\lambda$-GRPO & 82.80\% & 40.00\% & 21.69\% & 75.83\% & 1.62\% & \textbf{2.90\%} & 3.89\% & 32.68\% \\
\rowcolor{gray!30}\textbf{Ours} & \textbf{85.40\%} & \textbf{46.67\%} & \textbf{23.90\%} & 77.50\% & 1.50\% & 2.72\% & 4.22\% & \textbf{34.56\%} \\
\midrule
\multicolumn{9}{l}{\textit{Qwen3-8B}} \\
Base        & 87.40\% & 41.67\% & 28.68\% & 75.83\% & 2.89\% & 3.10\% & 2.89\% & 34.64\% \\
GRPO        & 89.20\% & 35.00\% & 28.68\% & 84.17\% & 3.03\% & 2.98\% & 3.24\% & 35.19\% \\
PPO         & 86.40\% & 36.67\% & 30.51\% & 80.00\% & 2.82\% & 2.17\% & 3.29\% & 34.55\% \\
Reinforce++ & 89.00\% & 45.00\% & 27.21\% & 84.17\% & 3.19\% & \textbf{4.35\%} & 3.29\% & 36.60\% \\
Dr.GRPO     & 87.80\% & 45.00\% & 27.94\% & 85.83\% & 2.50\% & 3.99\% & 3.10\% & 36.59\% \\
KTAE   & 86.00\% & 38.33\% & 29.41\%      & 78.33\% & 3.17\% & 3.99\%    & 2.64\% & 34.55\% \\
$\lambda$-GRPO & 90.00\% & \textbf{53.33\%} & 31.62\% & 80.00\% & 2.77\% & \textbf{4.35\%} & 3.57\% & 37.95\% \\
\rowcolor{gray!30}\textbf{Ours} & \textbf{90.20\%} & 46.67\% & \textbf{31.99\%} & \textbf{86.67\%} & \textbf{3.21\%} & \textbf{4.35\%} & \textbf{3.89\%} & \textbf{38.14\%} \\
\bottomrule
\end{tabular}
\caption{Accuracy (\%) on train-time optimization with rule-based rewards across models of different scales. Our method achieves consistent improvements under both Qwen3-1.7B and Qwen3-8B models.}
\label{tab:small-model}
}
\end{table*}

As shown in Table~\ref{tab:bottleneck}, we compare different information bottleneck designs across tasks and obtain three takeaways:
(1) A single-layer bottleneck delivers the best overall performance, with 4.17\% and 4.45\% accuracy on GPQA and HLE, indicating strong noise-filtering capability.
(2) A two-layer bottleneck excels on high-complexity math tasks but underperforms on others, likely due to overfitting to noise. However, for language models like Qwen, which are pretrained on rich mathematical data, deeper value structures may facilitate deeper understanding, yielding benefits in math domains. 
(3) No bottleneck yields modest gains on low-noise datasets but collapses on noisy ones (e.g., AMC23 drops to 50.83\%), showing high susceptibility to noisy or uncertain rewards.
Overall, moderate architectural bottlenecks substantially improve the robustness and cross-task generalization of value modeling under noisy RL Training.

\paragraph{VRPO’s two components interact synergistically rather than redundantly}
To examine whether the two core components of VRPO contribute redundantly or synergistically, we conducted a component-level ablation study. The results are reported in Table~\ref{tab:partialactivation} and Table~\ref{tab:bottleneck}, covering three representative settings: semantic-only training, IB-only training, and the full VRPO method that integrates both components.

Under the zero-bottleneck configuration, the model disables the information bottleneck and retains only the entropy/perplexity-based semantic loss, forming a semantic-only condition. This variant already outperforms the cold-start baseline on most metrics, indicating that semantic guidance alone can improve robustness under noisy supervision. 
Conversely, when using a 0\% token activation ratio, the semantic loss is effectively removed while the information bottleneck remains active, resulting in an IB-only condition. This setting also yields consistent improvements over the base model.

The fact that both isolated variants surpass the cold-start baseline demonstrates that each component independently enhances reasoning robustness. More importantly, the full VRPO configuration, which jointly combines semantic regularization with the information bottleneck, consistently achieves the best overall performance across both mathematical and scientific benchmarks. These results indicate that the two components interact synergistically rather than redundantly, jointly contributing to more stable and reliable advantage estimation.

\paragraph{VRPO remains effective across models of different scales} Table~\ref{tab:small-model} presents the results of the Qwen3-1.7B model on mathematical and scientific tasks under rule-based reward supervision during training-time optimization. The results demonstrate that, under noisy supervision, our method consistently outperforms baseline approaches such as PPO and GRPO, achieving an average accuracy of 58.37\% on mathematical tasks and 2.81\% on scientific tasks. These findings confirm the effectiveness of VRPO across model scales, and further highlight its robustness and generalization ability even in small-model settings.  

\begin{table}[ht]
\centering
\small
\resizebox{\columnwidth}{!}{
\begin{tabular}{lcccc|c}
\toprule
\makecell{\textbf{Activ-} \\ \textbf{ation}} &
\makecell{\textbf{MATH} \\ \textbf{500}} &
\makecell{\textbf{AIME} \\ \textbf{24}} &
\makecell{\textbf{Minerva} \\ \textbf{Math}} &
\makecell{\textbf{AMC} \\ \textbf{23}} &
\textbf{AVG} \\
\midrule
0\% & 72.20\% & 10.00\% & 20.22\% & 51.67\% & 38.52\% \\
20\% & 75.00\% & 6.67\% & 18.38\% & 54.17\% & 38.56\% \\
50\% & 73.40\% & 13.33\% & 19.49\% & 55.83\% & 40.51\% \\
70\% & 72.80\% & 10.00\% & 20.22\% & 58.33\% & 40.33\% \\
75\% & 73.20\% & 13.33\% & 19.49\% & 59.17\% & 41.29\% \\
\rowcolor{gray!30}\textbf{80\% (Ours)} & \textbf{74.40\%} & 13.33\% & 20.22\% & \textbf{61.67\%} & \textbf{42.41\%} \\
85\% & 73.60\% & 13.33\% & \textbf{23.90\%} & 58.33\% & 42.29\% \\
90\% & 71.00\% & 13.33\% & 21.69\% & 59.17\% & 41.29\% \\
100\% & 73.80\% & \textbf{16.67\%} & 19.12\% & 55.00\% & 41.15\% \\
\bottomrule
\end{tabular}
}
\caption{Impact of partial token activation on math reasoning performance using entropy/perplexity-based semantic loss. Activating approximately 80-85\% of tokens yields the best overall accuracy.}
\label{tab:partialactivation}
\end{table}
\begin{table}[ht]
\centering
\small
\resizebox{\columnwidth}{!}{
\begin{tabular}{lcccc|c}
\toprule
\makecell{\textbf{Activ-} \\ \textbf{ation}} &
\makecell{\textbf{MATH} \\ \textbf{500}} &
\makecell{\textbf{AIME} \\ \textbf{24}} &
\makecell{\textbf{Minerva} \\ \textbf{Math}} &
\makecell{\textbf{AMC} \\ \textbf{23}} &
\textbf{AVG} \\
\midrule
50\% & 70.40\% & 10.00\% & 19.85\% & 53.33\% & 38.40\% \\
\rowcolor{gray!30}\textbf{80\%} & 72.80\% & \textbf{16.67\%} & \textbf{20.96\%} & 55.00\% & \textbf{41.36\%} \\
100\% & \textbf{73.40\%} & 6.67\% & 18.01\% & \textbf{61.67\%} & 39.94\% \\
\bottomrule
\end{tabular}
}
\caption{Impact of partial token activation when entropy/perplexity scores are computed by an external model (Qwen-2.5-7B-Instruct). The optimal activation region remains consistent with the original setting.}
\label{tab:externalactivation}
\end{table}
\paragraph{Partial Token Activation Ratio Analysis and Self-Coupling Concern Verification}
To study how partial token activation influences semantic learning, we conducted an extensive activation percentage sweep on  mathematical reasoning tasks. The results are summarized in Table~\ref{tab:partialactivation}.

We observe that optimal performance does not come from a single sharply tuned threshold, but from a broader ~80–85\% activation interval, indicating the phenomenon is a robust region rather than a thresholding artifact. Performance varies smoothly outside this interval and remains relatively stable. 

To further address potential concerns regarding self-coupling, where entropy or perplexity is estimated by models at different stages of the same training system, we conducted an additional experiment using an external language model (Qwen-2.5-7B-Instruct) to compute token-level uncertainty scores. As shown in Table~\ref{tab:externalactivation}, the same activation pattern emerges: activating approximately 80\% tokens again achieves the best average performance.
The consistency of this trend across both internal and external uncertainty estimators supports the interpretation that the observed improvement reflects a generalizable property of partial token activation, rather than a model-specific artifact or self-induced coupling effect.

\paragraph{Improved Key Words Attribution in Advantage Estimation with VRPO}
\begin{table*}[ht]
\centering
\small
\begin{tabular}{lccccccc|c}
\toprule
\textbf{Domain} & \multicolumn{4}{c}{\textbf{Math}} & \textbf{Factuality} & \textbf{Science} & \textbf{Knowledge}& \textbf{ALL}    \\
\cmidrule(lr){1-1} \cmidrule(lr){2-5} \cmidrule(lr){6-8}
\textbf{Method} & \textbf{MATH500} & \textbf{AIME24} & \textbf{Minerva-Math} & \textbf{AMC23} & 
\textbf{SampleQA} & \textbf{GPQA} & \textbf{HLE} & \textbf{AVG} \\
\midrule
PPO &
21.10\% &
22.83\% &
27.70\% &
19.13\% &
\textbf{30.80\%}&
24.55\% &
\textbf{22.00\% }&
24.02\% \\
\rowcolor{gray!30}
\textbf{Ours} &
\textbf{27.30\%} &
\textbf{30.33\%} &
\textbf{30.80\%} &
\textbf{26.25\%} &
26.20\% &
\textbf{26.36\%} &
20.20\% &
\textbf{26.78\%} \\
\bottomrule
\end{tabular}
\caption{Top 20 Key Token Overlap (KTO) comparison between PPO and VRPO under noisy test-time optimization. Higher values indicate stronger alignment between high-advantage tokens and semantically meaningful keywords.}
\label{tab:keytoken}
\end{table*}

To move beyond qualitative visualizations and provide a more quantitative assessment of token-level credit assignment, we conduct an additional analysis comparing how different methods identify key tokens when estimating advantages under noisy training conditions. Our goal is to evaluate whether VRPO’s advantage estimates better align with semantically meaningful tokens in the output text.

As an approximate human-interpretable reference, we employ GPT-4o to extract salient keywords from each generated answer, which serve as a proxy for ground-truth key tokens. For each model, we then identify the top-20 tokens with the highest estimated advantages and compute their overlap with the extracted keywords. We report the \textbf{Key Token Overlap (KTO)} score, defined as:
\begin{equation}
\small
\text{KTO} = \frac{|T_{\text{model}} \cap T_{\text{gold}}|}{|T_{\text{gold}}|},
\end{equation}
where $T_{\text{model}}$ denotes the set of top-advantage tokens produced by the model and $T_{\text{gold}}$ denotes the keyword set extracted by GPT-4o.

We evaluate this metric on both mathematical and scientific reasoning benchmarks. Specifically, we sample 50 instances per dataset, except for AIME24 (30 samples) and AMC23 (40 samples), where all available data are used. Due to computational constraints, we focus on a direct comparison between PPO and VRPO under noisy train-time optimization settings.

The results in Table~\ref{tab:keytoken} show that VRPO achieves higher KTO scores than PPO, yielding an average improvement of over 2.76 points. This indicates that VRPO’s advantage estimates are more strongly aligned with semantically meaningful tokens, suggesting more faithful credit assignment. We also observe that high-advantage tokens sometimes include punctuation (e.g., “?”, “!”, “.”), likely because such tokens carry high attention weights and mark semantic boundaries in transformer models. Since these tokens are not included in GPT-4o’s keyword lists, they tend to lower absolute KTO values for all methods.

Overall, despite the heuristic nature of the reference signal, this quantitative analysis supports our core claim: by integrating semantic regularization with an information bottleneck in the value model, VRPO captures key tokens more effectively and produces more robust advantage signals, which in turn leads to improved training stability and performance under noisy supervision.

\paragraph{Statistical Significance and Random Seeds}
To validate the reliability of our results and ensure they are robust to initialization variance, we conducted additional experiments using multiple random seeds on the mathematical reasoning benchmarks with the Qwen3-8B model. The results are presented in Table~\ref{tab:random_seeds}.

\begin{table}[ht]
\centering
\small
\setlength{\tabcolsep}{1mm}{
\begin{tabular}{lcccc|c}
\toprule
\textbf{Seed} &
\makecell{\textbf{MATH} \\ \textbf{500}} & 
\makecell{\textbf{AIME} \\ \textbf{24}} & 
\makecell{\textbf{Minerva} \\ \textbf{Math}} & 
\makecell{\textbf{AMC} \\ \textbf{23}} & 
\textbf{AVG} \\
\midrule
22 & 89.80\% & \textbf{50.00\%} & 30.15\% & 84.17\% & 63.53\% \\
32 & \textbf{91.20\%} & 46.67\% & 30.15\% & 85.83\% & 63.46\% \\
\rowcolor{gray!30}\textbf{42 (Ours)} & 90.20\% & 46.67\% & \textbf{31.99\%} & \textbf{86.67\%} & 63.88\% \\
52 & 90.80\% & 46.67\% & 31.62\% & \textbf{86.67\%} & \textbf{63.94\%} \\
\bottomrule
\end{tabular}
\caption{Performance of VRPO on mathematical reasoning benchmarks across multiple random seeds using Qwen3-8B. The variance across independent runs is exceptionally low.}
\label{tab:random_seeds}
}
\end{table}

Our results indicate that the performance improvements reported in our main experiments are statistically significant and exceed normal training noise. The variance observed across these runs is exceptionally low, with the deviation of the final average accuracy remaining within 0.5\%. This confirms that the gains achieved by VRPO are stable and reproducible.

\subsection{Additional Sensitivity Analysis}
\begin{figure}[h]
    \centering
    \includegraphics[width= \linewidth]{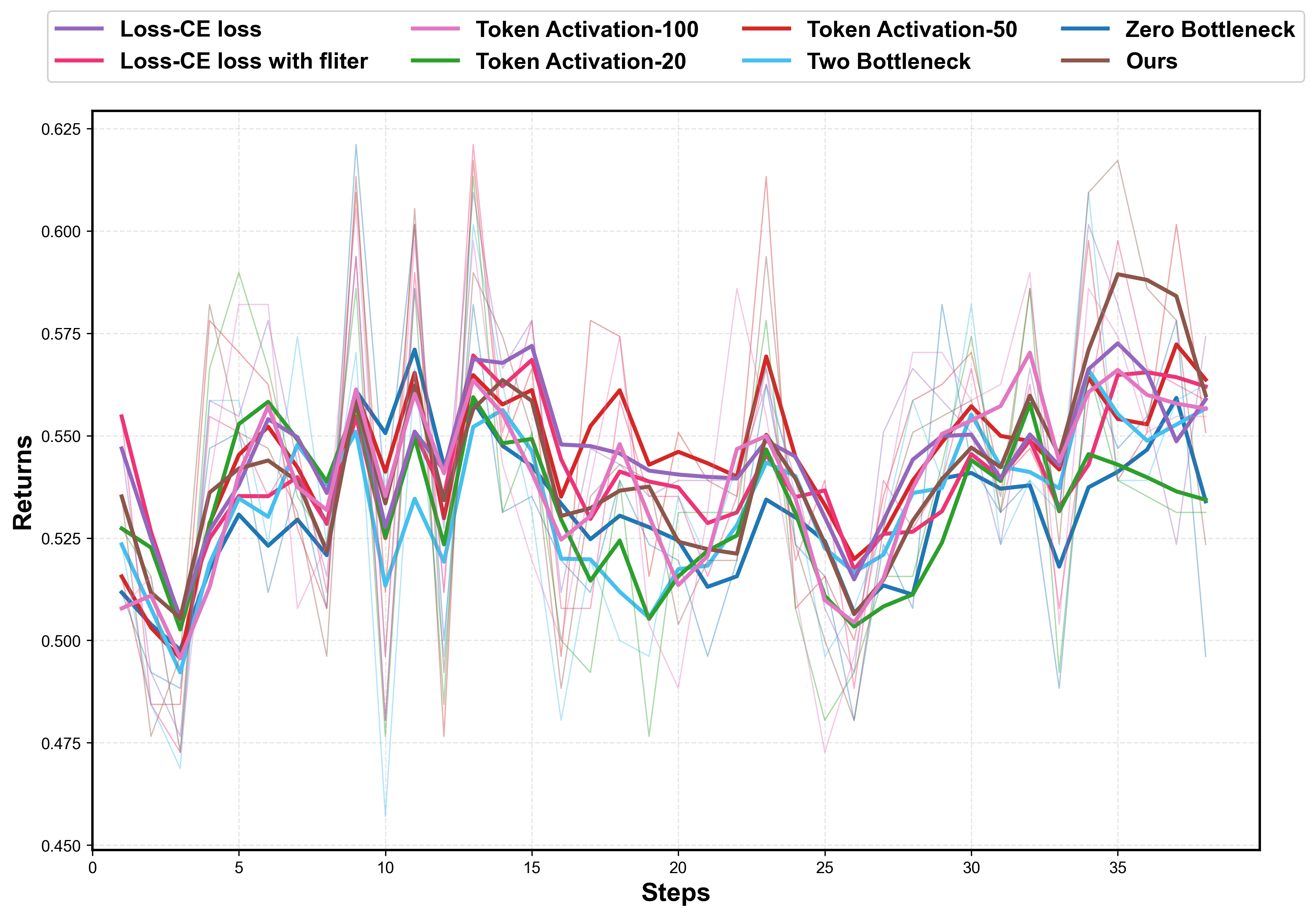}
    \caption{The evolution of training returns under test-time optimization settings on math datasets across different token-selection thresholds, learning strategies, and bottleneck configurations.}
    \label{fig:Sensitivity}
\end{figure}
We further analyze the training sensitivity of VRPO. Due to resource constraints, we take the experiments conducted under the settings of test-time optimization  on math datasets with varying token-selection thresholds, learning strategies, and bottleneck configurations as an example for analysis, as illustrated in Figure~\ref{fig:Sensitivity}.

Across all examined settings, the training curves exhibit smooth and low-variance behavior during the early stages, accompanied by a consistent upward trend. This indicates stable optimization dynamics and suggests that moderate variations in hyperparameters do not disrupt training. While the performance differences across configurations remain relatively small, our proposed setting achieves slightly higher returns throughout the training process with notably high returns observed particularly in the late stages of training.

These observations confirm that VRPO is not overly sensitive to hyperparameter tuning. Instead, its core components interact in a stable and well-behaved manner, providing mild yet consistent advantages across different configurations. This robustness further supports the practicality of VRPO in noisy reinforcement learning scenarios.

\section{Additional Visualization}

\begin{figure*}[h]
    \centering
    \includegraphics[width=0.75\textwidth]{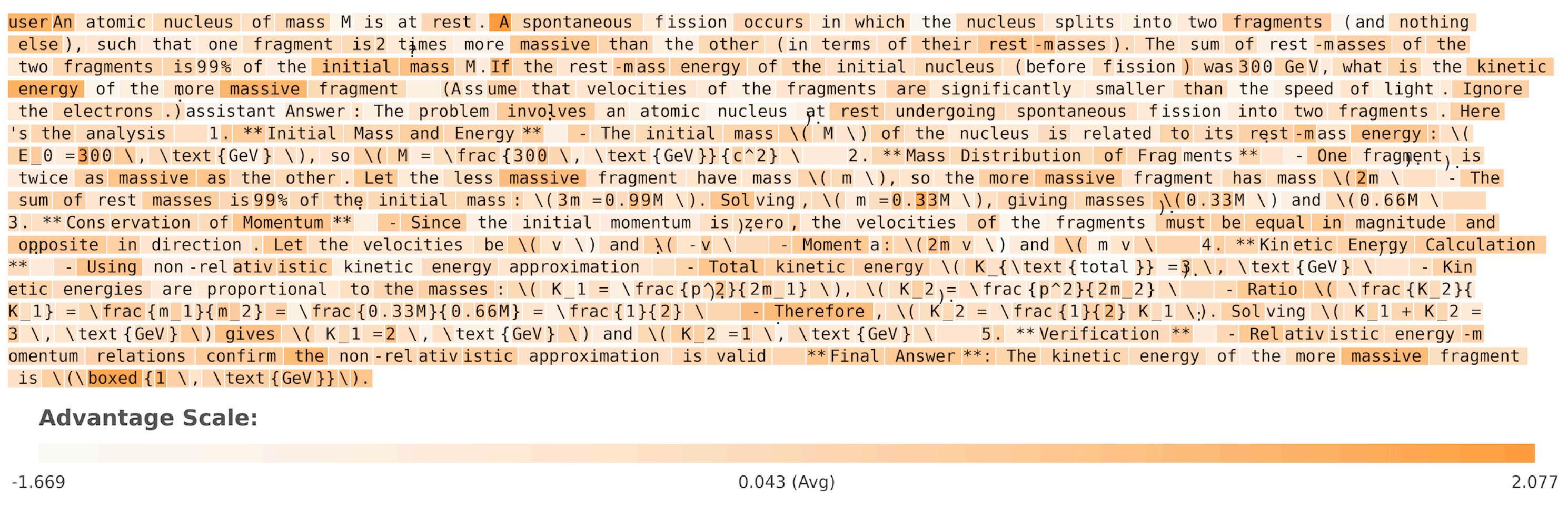}
    \caption{Our method on correct answer. Key information such as "initial mass M" is successfully captured.}
    \label{fig:ours-correct}
\end{figure*}

\begin{figure*}[h]
    \centering
    \includegraphics[width=0.75\textwidth]{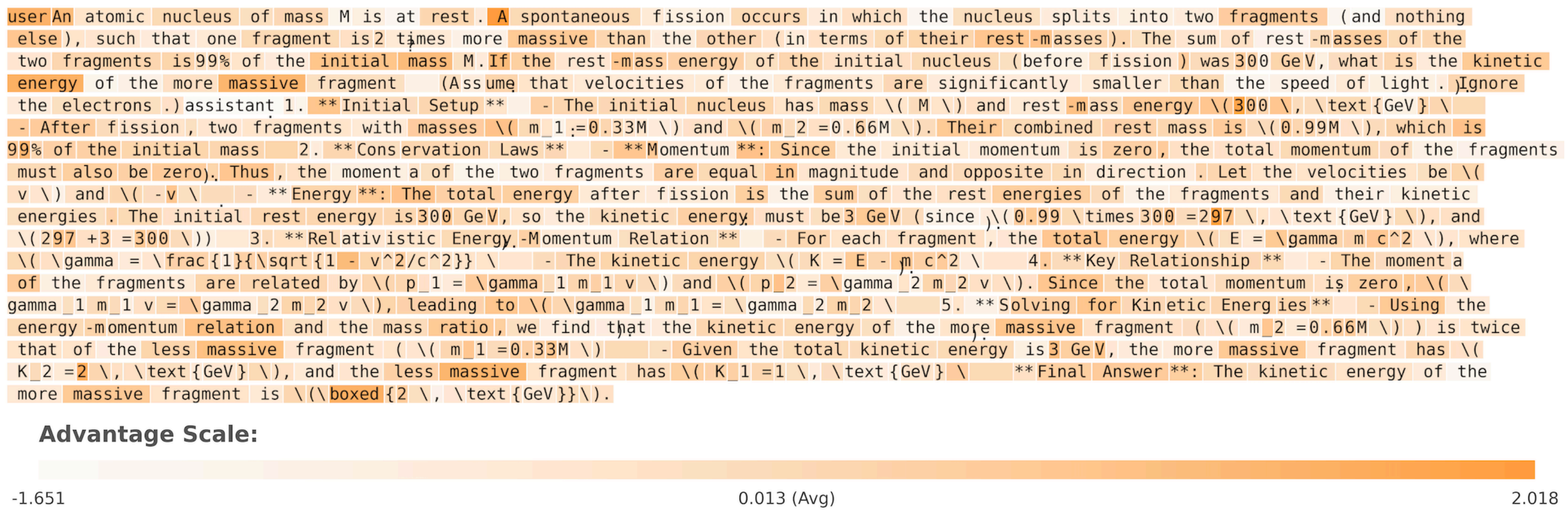}
    \caption{Our method on incorrect answer. The wrong prediction is appropriately assigned low advantage.}
    \label{fig:ours-wrong}
\end{figure*}

\begin{figure*}[h]
    \centering
    \includegraphics[width=0.75\textwidth]{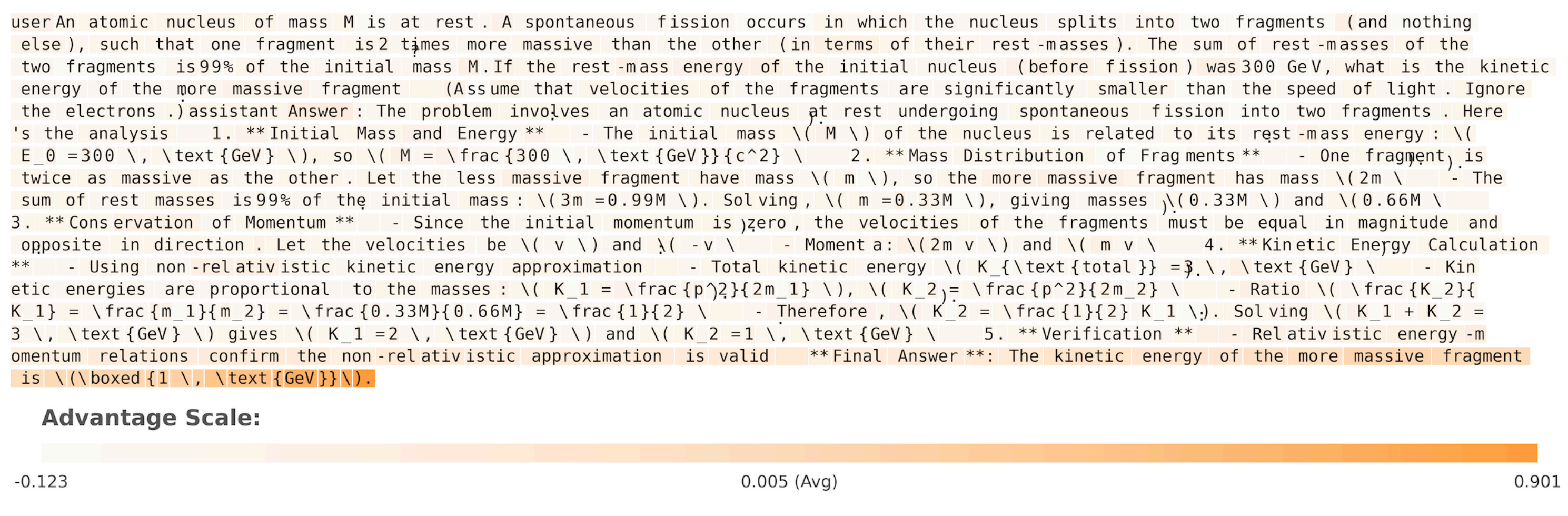}
    \caption{PPO on correct answer. The model fails to capture the semantic core and assigns low or inconsistent advantage.}
    \label{fig:ppo-correct}
\end{figure*}

\begin{figure*}[h]
    \centering
    \includegraphics[width=0.75\textwidth]{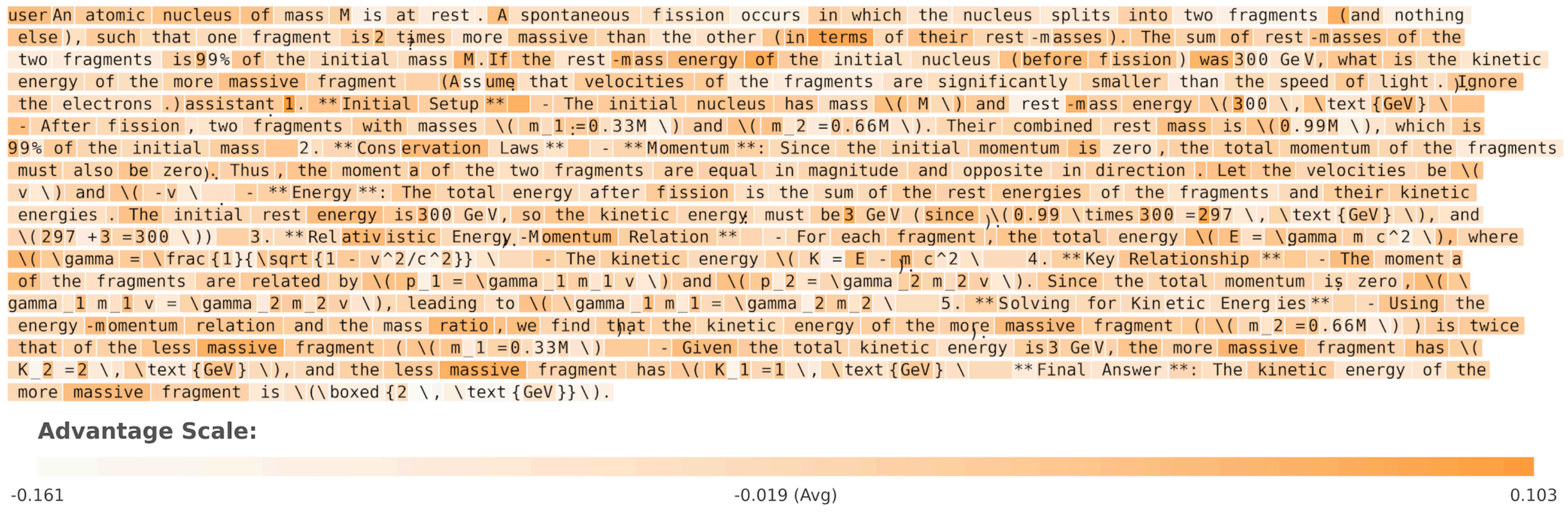}
    \caption{PPO on incorrect answer. High advantage is mistakenly assigned to both key and irrelevant content, weakening semantic discrimination.}
    \label{fig:ppo-wrong}
\end{figure*}

Figures~\ref{fig:ours-correct}, \ref{fig:ours-wrong}, \ref{fig:ppo-correct}, and \ref{fig:ppo-wrong} present a comparison of advantage estimations between PPO and our method (VRPO) on the same question outputs regardless of semantic quality. Furthermore, in the incorrect answer case (Figure~\ref{fig:example_fission_case}), under both correct and incorrect answer conditions.

Our method demonstrates a clear ability to capture key semantic components. In the correct answer case (Figure~\ref{fig:ours-correct}), it effectively identifies meaningful concepts such as "massive" and "initial mass M", while also showing a higher advantage estimate for the correct answer (Figure~\ref{fig:ours-correct}) compared to the incorrect one (Figure~\ref{fig:ours-wrong}). The model not only captures answer structures like box-based outputs but also makes more accurate judgments based on correctness.

In contrast, PPO fails to capture such distinctions. As seen in Figure~\ref{fig:ppo-correct}, the advantage estimation for the correct answer collapses, with later tokens receiving increasing advantage values. This pattern can lead to length hacking, where the model learns to generate longer outputs regardless of semantic quality. Furthermore, in the incorrect answer case (Figure~\ref{fig:ppo-wrong}), PPO assigns high advantages to both relevant and irrelevant content, indicating a lack of focus on key semantic cues and ultimately degrading training performance.

\begin{figure*}[!t]
\centering

\begin{tcolorbox}[
  colback=gray!4!white,
  colframe=gray!85!black,
  title=\textbf{System Prompt},
  fonttitle=\bfseries
]
\textbf{Role}: You are the AI call assistant of xxx Inc., capable of answering calls clearly and politely on behalf of the user and conversing with the caller. Based on the user's customized dialogue goals, you should first determine which goal applies to the conversation and respond accordingly. You do not know any personal information about the user beyond the task description and must inform the caller that any information they provide will be passed on to the user.

\textbf{Dialogue Goals (Customized)}:
\begin{itemize}
  \item \textbf{0: General Domain} – Used if the call doesn't match any specific domain. Be polite and professional. Try to answer or guide the caller and clarify their intent.
  \item \textbf{1: E-commerce/Retail} – If caller mentions down jacket performance, mention fast delivery, good warmth at -5°C, ask for cleaning tips, other styles, hotline, or exchange policy, inquire about washing temperature.
  \item \textbf{2: Technical Support} – If license issues arise for WX3510/WX5540, answer related questions, confirm email and phone, send manuals, ask for model and associated phone.
  \item \textbf{3: Financial Services – Pension} – If retirement certification reminder, say not remembered, will handle via WeChat, thank them, ask about past communication and WeChat face verification method.
  \item \textbf{4: Life Services} – If laundry fee update, respond with satisfaction, hope for discount, mention night service needs, ask about pricing explanation, training quality, and monthly discounts.
  \item \textbf{5: Financial Services – Trade Notice} – If transaction confirmation, say user unavailable. If medium-risk, ask for analyst opinion, arrange follow-up next day, inquire about transaction fees.
\end{itemize}
\end{tcolorbox}

\begin{tcolorbox}[
  colback=gray!4!white,
  colframe=gray!85!black,
  title=\textbf{Caller History and New Input},
  fonttitle=\bfseries,
  boxrule=1pt,
  top=3mm,
  bottom=3mm
]
\textbf{Conversation History}: \textit{(Empty)}

\textbf{Caller New Input}: Hello, this is the Forex Trading Customer Service Center. We are confirming that the currency pair you traded is USD/SGD, and the rate information is ready.
\end{tcolorbox}

\begin{tcolorbox}[
  colback=teal!2!white, 
  colframe=teal!80!black, 
  coltitle=white,
  title=\textbf{Model Response},
  fonttitle=\bfseries,
]
\textbf{Target Domain}: \textit{5: Financial Services – Trade Notice}

\textbf{Dialogue State}: \textit{Ongoing}

\textbf{Response}: Hello, this call is currently being handled by the smart assistant. The user is temporarily unavailable. If this is a medium-risk transaction, could we get advice from a market analyst? Please contact us again tomorrow for confirmation.
\end{tcolorbox}

\caption{A multi-turn example from the Honor-Dialogue dataset. The dataset features realistic, task-oriented, multi-domain conversations, in which each model response includes structured states. This example represents the situation of the first conversation.}
\label{fig:dialogue-example}
\end{figure*}

\begin{figure*}[!t]  
\centering

\begin{tcolorbox}[
  colback=gray!4!white,
  colframe=gray!85!black,
  title=\textbf{Core Prompt for Dialogue Logicality},
  fonttitle=\bfseries,
  boxrule=1pt,
  top=3mm,
  bottom=3mm
]
When scoring the intelligent call assistant's responses, please follow these steps:

1.Read and understand the entire call content.

2.Score from the dimension of dialogue logic rationality. The definition of dialogue logic rationality is: whether all responses provided by the assistant during the entire dialogue are clear, coherent, and effectively convey relevant information. A reasonable response should ensure the consistency, coherence, and relevance of information. Use a 1-5 scoring scale with specific criteria as follows:
\begin{itemize}
    \item \textbf{1 point:} The response exhibits severe logical inconsistencies and violates fundamental reasoning principles. The intent of the assistant is unclear, and the response fails to convey meaningful or usable information.

    \item \textbf{2 points:} The response contains multiple inconsistencies or conflicting statements. Although partially related to the query, it lacks sufficient contextual grounding and coherent reasoning, resulting in poor interpretability and weak informational value.

    \item \textbf{3 points:} The response is generally coherent but contains minor logical gaps or discontinuities that affect fluency. While relevant to the query, the explanation lacks depth or clarity in parts, limiting the overall effectiveness of communication.

    \item \textbf{4 points:} The response is logically consistent and well-structured, with clear and relevant content. Information is conveyed effectively and aligns well with the user’s intent, enabling smooth and coherent interaction.

    \item \textbf{5 points:} The response demonstrates strong logical consistency, clear reasoning, and precise information delivery. It not only addresses the query accurately but also facilitates deeper understanding, effectively guiding the conversation and enhancing overall interaction quality.
\end{itemize}

3.Based on the above scoring criteria and combined with the entire dialogue content, give a reasonable score, along with the scoring thinking process, deduction points, and modification suggestions to improve the score without changing the original meaning of the text.
\end{tcolorbox}

\caption{A core prompt for dialogue logicality assessment from the constructed rubric evaluation method. The rubric features clear scoring steps, explicit definition of logical rationality, and detailed 1-5 point end-point criteria, in which each evaluation requires supplementary scoring reasoning, deduction explanations, and optimization suggestions. }
\label{fig:scoring-guidelines}
\end{figure*}

\begin{figure*}[!t]
\centering

\begin{tcolorbox}[
width=1\textwidth,
  colback=gray!4!white,
  colframe=gray!85!black,
  coltitle=white,
  title=\textbf{Physics Fission Problem},
  fonttitle=\bfseries
]
\textbf{Problem}: An atomic nucleus of mass \( M \) is at rest. A spontaneous fission occurs in which the nucleus splits into two fragments (and nothing else), such that one fragment is 2 times more massive than the other (in terms of their rest-masses). The sum of rest-masses of the two fragments is 99\% of the initial mass \( M \). If the rest-mass energy of the initial nucleus (before fission) was 300 GeV, what is the kinetic energy of the more massive fragment?

\textbf{Assumptions}: Velocities of the fragments are significantly smaller than the speed of light. Ignore the electrons.
\end{tcolorbox}


\caption{Example problem from the GPQA dataset }
\label{fig:example_fission_case}
\end{figure*}

\end{document}